\pdfoutput=1

\documentclass[11pt]{article}

\usepackage[]{EMNLP2022}

\usepackage{times}
\usepackage{latexsym}

\usepackage[T1]{fontenc}

\usepackage[utf8]{inputenc}
\usepackage{microtype}

\usepackage{inconsolata}
\usepackage{graphicx}

\usepackage{booktabs}
\usepackage{amsmath}
\usepackage[capitalise]{cleveref}
\usepackage{makecell}
\usepackage{multirow}
\usepackage[frozencache,cachedir=minted-cache,]{minted}
\usepackage{hyperref}

\def\Snospace~{\S{}} 

\newcommand{\modeldetail}[1]{

\begin{table*}[#1]
\centering
\footnotesize
\resizebox{\textwidth}{!}{%
\begin{tabular}{
    l@{\hskip 0.5in}
    c
    c
    c
    c
    c
    c
    c
}
\toprule 
Model       &  Type & Pretraining Objective & Context Size & First Position & \# Layers & Hidden Size & \# Params    \\
\midrule
\multicolumn{8}{c}{RoBERTa family \citep{Liu2019:RoBERTa}} \\
\midrule
$\text{RoBERTa}_\text{BASE}$ & encoder-only & Masked Language Modeling & 514 & 2 & 12 & 768 & 123M \\
$\text{RoBERTa}_\text{LARGE}$ & encoder-only& Masked Language Modeling & 514 & 2 & 24 & 1024 & 325M \\

\midrule
\multicolumn{8}{c}{BART family \citep{Lewis2020:BART}} \\
\midrule

$\text{BART}_\text{BASE}$ & encoder-decoder& Masked Language Modeling & 1024 & 2 &6 & 768 & 140M \\
$\text{BART}_\text{LARGE}$ & encoder-decoder& Masked Language Modeling & 1024 & 2 &12& 1024 & 400M \\

\midrule
\multicolumn{8}{c}{GPT2 family \citep{Radford2019:GPT2}} \\
\midrule

$\text{GPT2}$ & decoder-only & Next Token Prediction &1024 & 0&  12 & 768 & 125M \\
$\text{GPT2}_\text{MEDIUM}$ & decoder-only & Next Token Prediction & 1024 & 0 & 24 & 1024 & 345M \\

\midrule
\multicolumn{8}{c}{OPT family \citep{Zhang2022:OPT}} \\
\midrule

$\text{OPT}_\text{125M}$ & decoder-only & Next Token Prediction& 2048 & 2 & 12 & 768 & 125M \\
$\text{OPT}_\text{350M}$ &decoder-only & Next Token Prediction& 2048  & 2 & 24& 1024& 350M \\
$\text{OPT}_\text{2.7M}$ &decoder-only & Next Token Prediction& 2048 & 2 & 32 & 2560 & 2.7B \\
$\text{OPT}_\text{13B}$ &decoder-only &Next Token Prediction&  2048 & 2 & 40 & 5120 & 13B \\
$\text{OPT}_\text{30B}$ &decoder-only &Next Token Prediction&  2048 & 2 & 48 & 7168 & 30B \\

\bottomrule
\end{tabular}
}
\caption{Details of the models we used in this paper.}
\label{tab:model_detail}
\end{table*}{}

}
\newcommand{\examplecode}[1]{

}
\newcommand{\datasetdetail}[1]{

\begin{table}[#1]
\centering
\footnotesize
\begin{tabular}{
    l@{\hskip 0.6in}
    c
    c
}
\toprule 
Dataset       &  \# Train  & \# Test/Validation  \\
\midrule
BliMP & - & 67000 \\
COPA & 400 & 100 \\
PIQA & 16113 & 1838 \\
WinoGrande & 40398 & 1267 \\
ARC (Easy) & 2251 & 2376 \\
MRPC & 3668 & 408 \\
RTE & 2490 & 277 \\
CoLA & 8551 & 1043 \\

\bottomrule
\end{tabular}
\caption{Dataset statistics we used in this work.}
\label{tab:dataset_detail}
\end{table}{}

}
\newcommand{\fthyperparam}[1]{

\begin{table}[#1]
\centering
\footnotesize
\begin{tabular}{
    l@{\hskip 0.2in}
    r
}
\toprule 
Parameter       &  Value  \\
\midrule
Learning rate & $\{0.0001, 0.0002, 0.0003\}$ \\
Batch size & $\{16, 32\}$ \\
\# Train Epochs & 10 \\
Early Stopping & On \\
Early Stopping Tolerance & 3 \\
Optimizer & AdamW \\
Learning Rate Schedule & Linear \\
Weight Decay & 0.0 \\
Warm Up & 6\% of initial training steps \\

\bottomrule
\end{tabular}
\caption{Summary of hyperparamters used in finetuning experiments.}
\label{tab:finetune_hyperparam}
\end{table}{}

}
\newcommand{\modelposenc}[1]{

\begin{table*}[#1]
\centering
\footnotesize
\begin{tabular}{
    l
    c
    c
}
\toprule 
Name       &  Release Year  & Positional Encoding Type  \\
\midrule
BERT \citep{Devlin2019:BERT} & 2019 & Learned Absolute \\
RoBERTa \citep{Liu2019:RoBERTa} & 2019 & Learned Absolute \\
GPT2 \citep{Radford2019:GPT2} & 2019 & Learned Absolute \\
BART \citep{Lewis2020:BART} & 2020 & Learned Absolute \\
LongFormer \citep{longformer} & 2020 & Learned Absolute \\
T5 \citep{Raffel2020:T5} & 2020 & Relative Learned Bias \\
GPT3 \citep{Brown2020:GPT3} & 2020 & Learned Absolute \\
GPT-Neo \citep{gpt-neo} & 2021 & Learned Absolute \\
Fairseq-Dense \citep{fairseq} & 2021 & Fixed Absolute  \\
ShortFormer \citep{shortformer} & 2021 & Fixed Absolute \\
GPT-J \citep{mesh-transformer-jax} & 2021 & Rotary \\
GPT-NeoX \citep{Black2022:GPTNeoX} & 2022 & Rotary \\
OPT \citep{Zhang2022:OPT} & 2022 & Learned Absolute \\
PaLM \citep{palm} & 2022 & Rotary \\

\bottomrule
\end{tabular}
\caption{Positional encoding of commonly used pretrained language models.}
\label{tab:models_pe}
\end{table*}{}

}
\newcommand{\sweepresults}[1]{

\begin{table*}[#1]
\centering
\footnotesize

\resizebox{\textwidth}{!}{%
\begin{tabular}{
    l@{\hskip 0.1in}
    c
    c@{\hskip 0.4in}
    c
    c@{\hskip 0.4in}
    c
    c@{\hskip 0.4in}
    c
    c
}
\toprule 
 & \multicolumn{8}{c}{Phase shifts} \\
 & \multicolumn{2}{c}{$k=0$} & \multicolumn{2}{c}{$k=100$} & \multicolumn{2}{c}{$k=200$} & \multicolumn{2}{c}{$k=300$} \\
 \cmidrule{2-9}
Model & Learning Rate & Batch Size & Learning Rate & Batch Size & Learning Rate & Batch Size & Learning Rate & Batch Size \\
\midrule
\multicolumn{9}{c}{CoLA} \\
\midrule

$\text{RoBERTa}_\text{BASE}$ & 
0.00002 & 
32 & 

0.00002 & 
16 & 

0.00002 & 
16 & 

0.00002 & 
16 

\\ 

$\text{RoBERTa}_\text{LARGE}$ & 
0.00003 & 
32 & 

0.00003 & 
32 & 

0.00001 & 
32 & 

0.00002 & 
16 
\\ 

$\text{BART}_\text{BASE}$ & 
0.00002 & 
32 & 

0.00003 & 
16 & 

0.00002 & 
16 & 

0.00002 & 
32 

\\ 

$\text{BART}_\text{LARGE}$ & 
0.00002 & 
16 & 

0.00003 & 
32 & 

0.00003 & 
16 & 

0.00003 & 
32 

\\ 

GPT2 & 
0.00002&16 & 0.00003&32 & 0.00003&16 & 0.00003&16 \\

$\text{GPT2}_\text{MEDIUM}$ & 
0.00002&32 & 0.00001&16 & 0.00003&16 & 0.00003&16 \\

$\text{OPT}_\text{125M}$ & 
0.00002&16 & 0.00001&16 & 0.00001&32 & 0.00001&16 \\

$\text{OPT}_\text{350M}$ & 
0.00001&16 & 0.00001&32 & 0.00002&32 & 0.00001&16 \\

\midrule
\multicolumn{9}{c}{MRPC} \\
\midrule

$\text{RoBERTa}_\text{BASE}$ & 
0.00002 & 
32 & 

0.00003 & 
16 & 

0.00003 & 
32 & 

0.00001 & 
32 
\\ 

$\text{RoBERTa}_\text{LARGE}$ & 
0.00002 & 
32 & 

0.00001 & 
16 & 

0.00002 & 
32 & 

0.00002 & 
16 
\\ 

$\text{BART}_\text{BASE}$ & 
0.00001 & 
16 & 

0.00003 & 
32 & 

0.00002 & 
16 & 

0.00003 & 
16 
\\ 

$\text{BART}_\text{LARGE}$ & 
0.00002 & 
16 & 

0.00003 & 
16 & 

0.00002 & 
16 & 

0.00003 & 
16 
\\ 

GPT2 & 
0.00002&16 & 0.00003&16 & 0.00002&16 & 0.00003&16 \\

$\text{GPT2}_\text{MEDIUM}$ & 
0.00002&16 & 0.00003&16 & 0.00003&16 & 0.00003&16 \\

$\text{OPT}_\text{125M}$ & 
0.00003&16 & 0.00002&32 & 0.00002&16 & 0.00003&32 \\

$\text{OPT}_\text{350M}$ & 
0.00003&32 & 0.00001&16 & 0.00001&32 & 0.00001&32 \\

\midrule
\multicolumn{9}{c}{RTE} \\
\midrule

$\text{RoBERTa}_\text{BASE}$ & 
0.00002 & 
16 & 

0.00003 & 
16 & 

0.00002 & 
16 & 

0.00002 & 
16 

\\ 

$\text{RoBERTa}_\text{LARGE}$ & 
0.00003 & 
32 & 

0.00001 & 
32 & 

0.00003 & 
32 & 

0.00001 & 
32 

\\ 

$\text{BART}_\text{BASE}$ & 
0.00003 & 
16 & 

0.00003 & 
32 & 

0.00002 & 
32 & 

0.00003 & 
16 

\\ 

$\text{BART}_\text{LARGE}$ & 
0.00003 & 
32 & 

0.00003 & 
16 & 

0.00002 & 
16 & 

0.00003 & 
16 

\\ 

GPT2 & 
0.00001&16 & 0.00003&16 & 0.00003&16 & 0.00003&16 \\

$\text{GPT2}_\text{MEDIUM}$ & 
0.00002&16 & 0.00003&16 & 0.00001&16 & 0.00002&32 \\

$\text{OPT}_\text{125M}$ & 
0.00003&16 & 0.00001&32 & 0.00001&16 & 0.00001&32 \\

$\text{OPT}_\text{350M}$ & 
0.00001&16 & 0.00001&16 & 0.00001&32 & 0.00001&16 \\

\bottomrule
\end{tabular}
}
\caption{Result of hyperparamter sweep for finetuning experiments.}
\label{tab:sweep_results}
\end{table*}{}

}

\title{The Curious Case of Absolute Position Embeddings}

\author{Koustuv Sinha$^\ddagger$$^{\dagger}$$^*$ \quad Amirhossein Kazemnejad $^\ddagger$$^*$ \vspace{0.2em} \\  {\bf Siva Reddy}$^\ddagger$ \quad {\bf Joelle Pineau}$^\dagger{}^\ddagger$ \quad {\bf Dieuwke Hupkes}$^\dagger$ \quad {\bf Adina Williams}$^\dagger$ \vspace{0.4em} \\
$^\ddagger$ McGill University / Mila - Quebec AI; $^\dagger$ Meta AI \\
  \texttt{\{koustuv.sinha,amirhossein.kazemnejad\}@mail.mcgill.ca} \\}

\begin{document}
\maketitle
\def\thefootnote{*}\footnotetext{Equal contributions.}\def\thefootnote{\arabic{footnote}}
\begin{abstract}
Transformer language models encode the notion of word order using positional information. 
Most commonly, this positional information is represented by absolute position embeddings (APEs), that are learned from the pretraining data. However, in natural language, it is not \emph{absolute} position that matters, but \emph{relative position}, and the extent to which APEs can capture this type of information has not been investigated.
In this work, we observe that models trained with APE over-rely on positional information to the point that they break-down when subjected to sentences with shifted position information.
Specifically, when models are subjected to sentences starting from a non-zero position (excluding the effect of priming), they exhibit noticeably degraded performance on zero- to full-shot tasks, across a range of model families and model sizes. 
Our findings raise questions about the efficacy of APEs to model the relativity of position information, and invite further introspection on the sentence and word order processing strategies employed by these models.



\end{abstract}

\section{Introduction}


Recently, Transformer \citep{vaswani2017} language models (TLMs) have been widely used for natural language applications. Such models incorporate positional encodings: vectors encoding information about the order of words in context. Many models, such as RoBERTa \cite{Liu2019:RoBERTa}, GPT3 \cite{Brown2020:GPT3} and OPT \cite{Zhang2022:OPT}, utilize \textit{absolute} position embeddings (APEs) that directly encode absolute (linear) word order. APEs appear to contribute to the performance of such models; although when they are removed, some models become sensitive to ablative word scrambles \citep{sinha-etal-2021-masked}, while others work optimally \citep{haviv2022}. Thus, what precisely APEs contribute remains unclear. 

It is conceivable that APEs may enable the model to handle the relative distances between words. 
If models were somehow learning relative position information despite using \emph{absolute} positional embeddings, we would expect sentence encodings to be the same in most cases, regardless of where they appear in the context window. For example, the meaning of ``smoking kills'' should be constant in ``Kim said \textit{smoking kills}'' (positions 2--3)  and ``It was commonly believed by most adult Americans in the 90s that \textit{smoking kills}'' (positions 13--14), despite the fact that these words appear in different absolute positions.
Given this, our central question is: do APEs enable the model to learn the relative distances between the words in a sentence? 

Prior work has attempted to explore the consequences of APEs using probing methods \cite{wang2021on}.
APEs have been found to not capture the meaning of absolute or relative positions \cite{wang-chen-2020-position}.
APEs have also been found to bias model output with positional artefacts \cite{luo-etal-2021-positional}, leading to better performance on token to position de-correlation \cite{ke2021}.
\citet{haviv2022} even find that causal TLMs perform adequately even without an explicit APEs.
However, a systematic study on relativity of positional encodings is still needed.

\begin{figure}
    \centering
    \resizebox{\linewidth}{!}{
    \includegraphics[width=\textwidth]{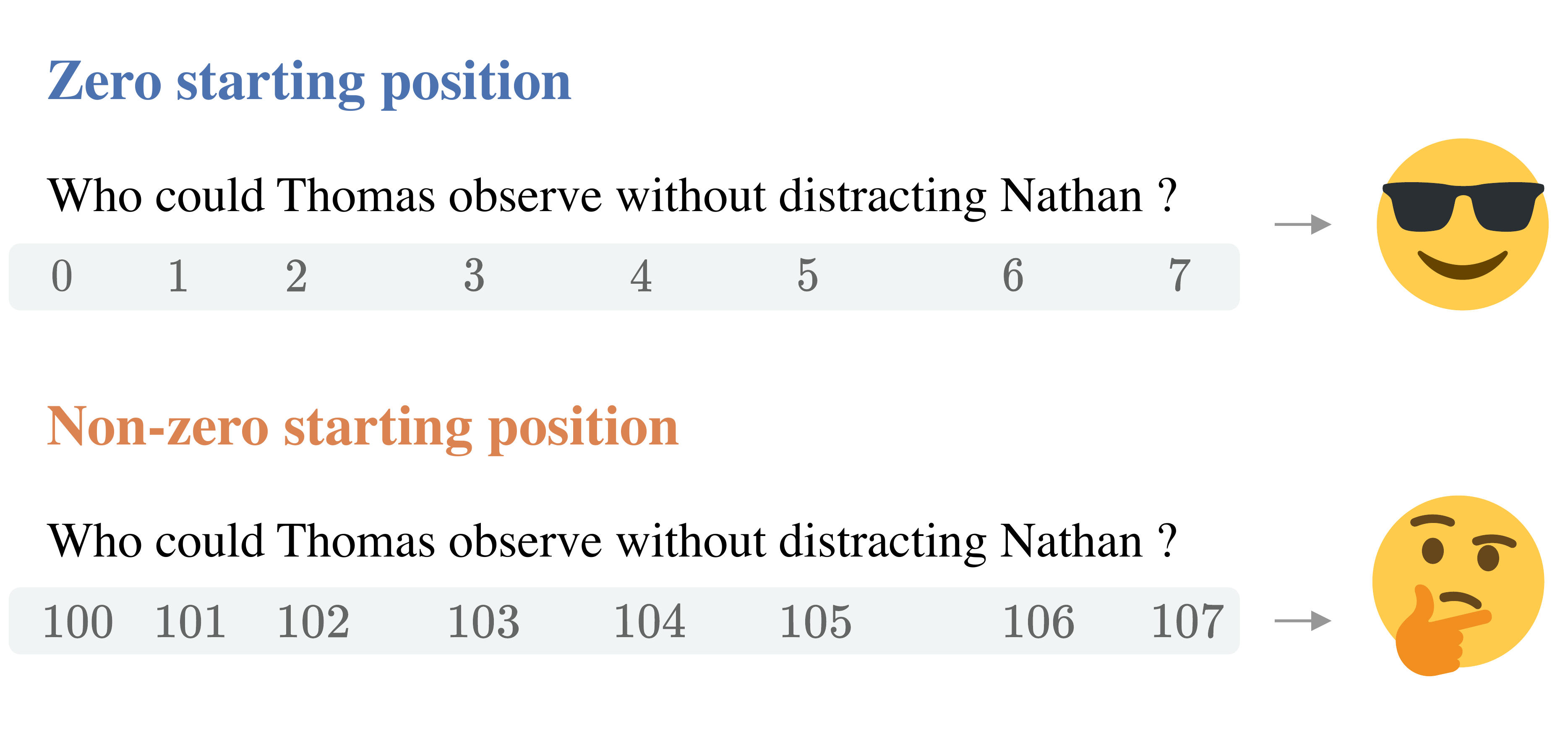}}
    \caption{Transformer models with absolute positional embeddings have different representations for sentences starting from non-zero positions. }
    \label{fig:bert_confused}
\end{figure}

To better understand the relativity of absolute position embeddings, we first need to ascertain the robustness of relative position understanding for a given input. 
TLMs are typically trained in a batch containing multiple sentences, with a limited sequence window size, which is typically much larger than an average sentence.
We hypothesize that a systematic model should encode the same sentence equally throughout this context window.
However, evaluating the encoding of a sentence starting from any position in this window in isolation is hard, as the representation of the sentence would depend on the prior context \cite{misra2020exploring,kassner-schutze-2020-negated}. 

In this work, we subject models from several different architectures and sizes to \textit{phase shifting}. 
In this paradigm, the sentences exposed to the model are provided contiguous position identifiers starting from a non-zero position (\autoref{fig:bert_confused}).
Such inspection allows us to gauge the model's sentence encodings on different positions, emulating sub-window sentence representation, while factoring out the influence of prior context.  
We investigate several zero shot, few shot and full shot tasks by shifting the start positions of the sentences. We observe the following:

\begin{itemize}
    \item TLMs display different sub-window sentence representation capabilities, resulting in decreased zero shot task performance and variability in sentence perplexities. 
    \item Autoregressive models, including the recently published OPT \cite{Zhang2022:OPT}, show erratic zero and few-shot performance on sub-window representations, highlighting the brittleness of in-context learning evaluation.
    \item Masked Language Models (MLMs) encode sentences in non-standard positions better than their autoregressive counterparts.
    \item During fine-tuning models suffer drastically on cross phase-shifted evaluation, suggesting position specific overfitting.
\end{itemize}

\noindent We aim to raise awareness about issues with APEs, which are still widely used in pre-training large language models. 
Our results highlight the severity of position shortcuts taken by the model during pre-training and fine-tuning, and imply that TLMs may have vastly varying sub-window sentence representation capability than previously assumed. 
We will release the code and analysis used in this work on Github. \footnote{\href{https://github.com/kazemnejad/lm_pos_investigations}{https://github.com/kazemnejad/lm\_pos\_investigations}}





\section{Approach}

Position encodings used by TLMs come in three broad categories: fixed sinusoidal embeddings as proposed by \citet{vaswani2017}, absolute or learned popularized by BERT \cite{Devlin2019:BERT} family of masked language models, and relative positions \citep{shaw-etal-2018-self} used by T5 \cite{Raffel2020:T5}.
\citet{wang2021on} presents a comprehensive overview of current encoding strategies.

Despite being an older method, absolute positional embeddings (APEs) are reportedly better than its relative counterparts on several tasks \citep{ravishankar-etal-2021-multilingual}, and are still used by majority of the large pre-trained TLMs, including the recently released OPT \cite{Zhang2022:OPT}. 
APEs compute token representation after adding the input token to the position embedding for the corresponding position: $x_i = \theta_{W}[w_i] + \theta_{P}[i]$,
where, $\theta_W \in \mathbf{R}^{|V| \times d}$ is the token vocabulary of size $|V|$, embedding dimension $d$, and the absolute position embedding matrix $\theta_P \in \mathbf{R}^{|T|\times d}$, where $T$ is the maximum context window size of the model. 
Now, a sentence $S=[w_1,w_2 ... w_n]$ containing $n$ tokens, is mapped during inference to positions 1,2, ... $n$ contiguously for all models.

TLMs offer various sizes of \textit{context window}, which is the maximum sequence length in tokens it can train and infer on. 
Since this context window is usually larger than the average sentence length, multiple sentences can be packed together to ``fill" the context window during pre-training.
This allows TLMs to learn that sentences can start from various positions in their context window. If models trained with APEs do encode relativity of position, then the sentence representations should be roughly equal throughout the context window, regardless of their starting position.

\subsection{Phase Shift Methodology}
\label{app:trigger_tokens}
\begin{figure}
    \centering
    \resizebox{\linewidth}{!}{
    \includegraphics[width=\textwidth]{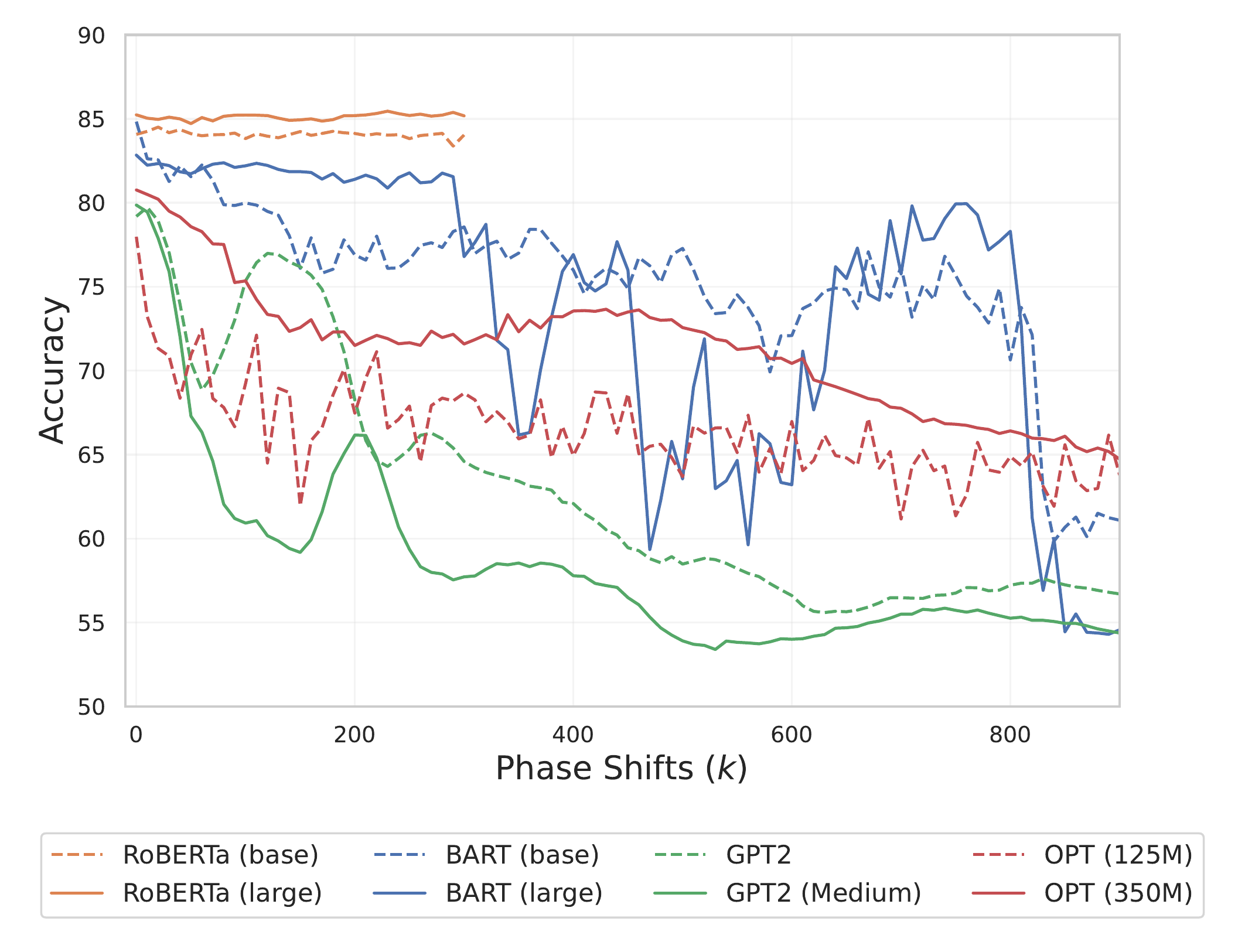}}
    \caption{Acceptability Scores in BLiMP \cite{warstadt-etal-2020-BLiMP-benchmark} dataset across different phase shifts. RoBERTa only supports context window of size $T=512$, so we capped the scores to phase shift $k=300$ to allow for sentences of maximum length in BLiMP to be evaluated.}
    \label{fig:acceptability}
\end{figure}

To understand the relativity of APEs, we examine the model performance under \textit{phase shift} conditions.
Phase shift\footnote{More related to our work, \citet{kiyono2021} train a Transformer model from scratch using shifted positional embeddings for machine translation, and observe improved performance in extrapolation and intrapolation setup.} involves right-shifting the absolute positions of all tokens in the sentence by an equal distance $k$, such that the tokens are now mapped to new positions $1+k, 2+k, ... , n+k$, or $x_i = \theta_{W}[w_i] + \theta_{P}[i+k]$.
As such, phase shifting changes only the absolute position, but preserves the relative distances between tokens in the a sentence.
Theoretically, we can shift the positions within the context window as long as $k+n \leq T$. 
For example, given phase shift $k=100$, and sentence length of $n$, we could have the following vector of position ids:

$$
\Vec{p} = [101, 102, 103, \dots, n+100]
$$

While computing the task scores and perplexities of the models, we observed that all of the models exhibit poor task performance on phase shifts.
Due to the non-shiftable nature of the \texttt{[CLS]} token in masked language models (MLMs), we first fix the position of \texttt{[CLS]} token to start position during phase shifting, which results in significantly improved performance for all models:

$$
\Vec{p} = [1, 102, 103, \dots, n+100]
$$

Futhermore, we observed yet another marked improvement in task performance when we use \textit{special tokens} in the beginning of the sentence: typically the end-of-sentence (\texttt{[EOS]}) token in case of MLM models (RoBERTa, BART).
An explanation for this ambiguity in results is that typically when models are pre-trained, multiple sentences are packed together in the context window by delimiting the start of each sentence with an \texttt{[EOS]} token
\footnote{While this is not the case for GPT2, we also observed improved performance in some cases when we add a beginning of sentence (\texttt{[BOS]}) token to the sentence and add a special \texttt{[EOS]} token to delimit the start of a sentence.}.
Thus, in all of our results, we opt with this configuration (adding an \texttt{[EOS]} token before the sentence) to ensure fairer evaluation for all model families.
Concretely, the input to a model uses the following template \footnote{In cases where a model does not have the \texttt{[CLS]} token, we instead use \texttt{[BOS]}.
If none of those are available, we replace it with \texttt{[EOS]} (so a total of two \texttt{[EOS]}'s will be prepended).}:
$$
\texttt{[CLS]}\texttt{[EOS]} \texttt{<sentence>}
$$

\begin{figure}
    \centering
    \resizebox{\linewidth}{!}{
    \includegraphics[width=\textwidth]{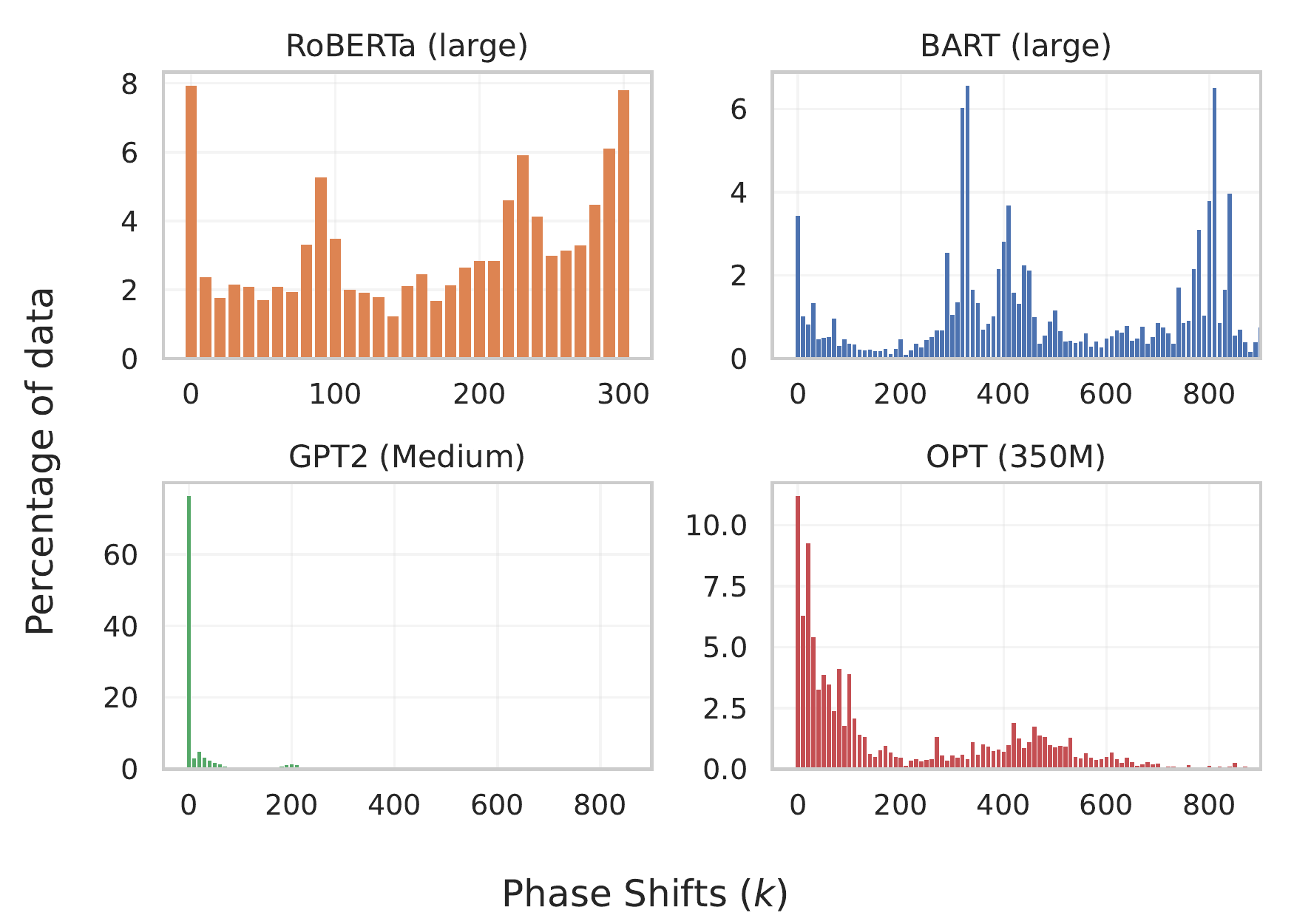}}
    \caption{Distribution of sentences in BLiMP \cite{warstadt-etal-2020-BLiMP-benchmark} having the lowest perplexities (i.e., are deemed most acceptable) for each phase shift.}
    \label{fig:ppl_density_main}
\end{figure}

\begin{figure*}
    \centering
    \resizebox{\linewidth}{!}{
    \includegraphics[width=0.9\textwidth]{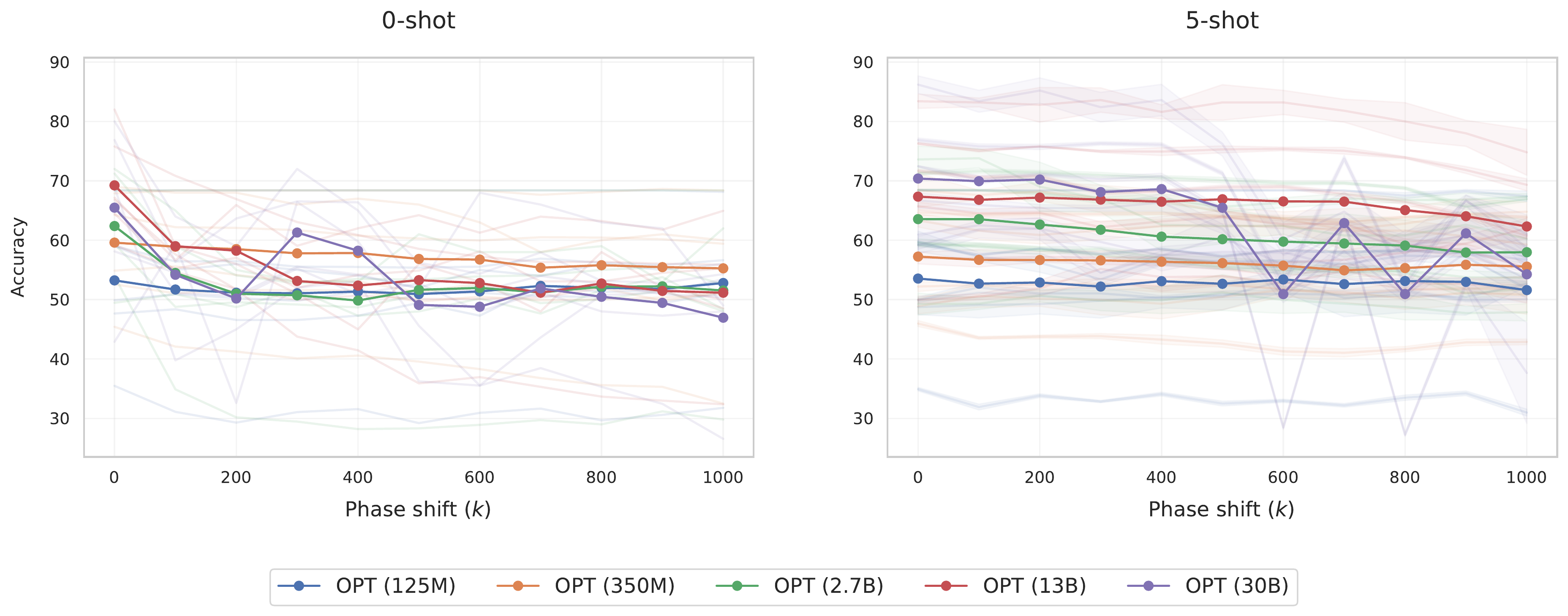}}
    \caption{Aggregate performance of OPT family on six NLP tasks when various phase shifts are applied.}
    \label{fig:prompt_ps_aggr}
\end{figure*}

\section{Impact of phase shifts on grammatical acceptability}
\label{sec:acceptability}

First, we investigate the impact of phase shifting on the model performance.
We compute the perplexities of several publicly available models---RoBERTa \citep{Liu2019:RoBERTa}, BART \citep{Lewis2020:BART}, GPT2 \citep{Radford2019:GPT2} and OPT \citep{Zhang2022:OPT}---to evaluate the grammatical acceptability capabilities of the model, using the BLiMP \cite{warstadt-etal-2020-BLiMP-benchmark} benchmark.\footnote{We adopt the perplexity computation strategy for RoBERTa and BART from \citet{salazar2019masked}} 
We compute the task score by comparing grammatical and ungrammatical sentence perplexities, and applying the phase shift in increasing values of $k$ to the sentences and models (\autoref{fig:acceptability}). 

We observe that the task performance of all models, except for RoBERTa, drastically suffers from phase shifting. Autoregressive models in particular display worse results. This is likely due to a mismatch of position information learned due to the causal language modelling objective vs the position information provided to the model during phase shift \cite{haviv2022}. We also compare the perplexities of each sentence across different phase shifts and plot the frequency of sentences having the lowest perplexity in each $k$ (\autoref{fig:ppl_density_main}). We observe in GPT2 that more than 70\% of the sentences have their best perplexity in $k=0$, highlighting a severe zero-position bias. $\text{OPT}_\text{350M}$ has better sub-window sentence representation capacity than similarly sized GPT2, which is also evident from the acceptability results in \autoref{fig:acceptability}.


\section{Impact of phase shifts on in-context learning}
\label{sec:prompting}



More recently, zero-shot and few-shot inference, commonly referred to as in-context learning, have become a de facto standard in evaluating pretrained language models \citep{Brown2020:GPT3}.
In this approach, the model's predictions are produced by conditioning it on certain prompts, such as instructions (zero-shot setting) or a few examples of input-output pairs (few-shot setup).
In both cases, the model faces an extended input text, and we suspect it will be affected by deficiencies of APE.
To evaluate this hypothesis, we employ an experimental setup similar to \autoref{sec:acceptability}.
Under zero-shot and five-shot inference regimes, we assess the model performance on standard NLP tasks when it is fed with inputs in increasing values of phase shifts. 
We choose OPT model family, because it is available in a wide range of sizes (125M to 30B parameters), allowing allows us to examine the behavior of APE at different scales.
Moreover, our evaluations take into account four tasks reported in the original paper: Winogrande \citep{Sakaguchi2020:WINOGRANDE}, COPA \citep{Gordon2012:COPA}, PIQA \citep{Bisk2020:PIQA}, and ARC \citep{Clark2018:ARC} as well as two classification datasets from GLUE benchmark \citep{Wang2018:GLUE}: MRPC and RTE.
We provide an aggregated view of the models' performance on all six accuracy-dominated benchmarks in \autoref{fig:prompt_ps_aggr}. The detailed plots for each task are in \autoref{app:detailed_prompts}.

\begin{figure}
    \centering
    \resizebox{\linewidth}{!}{
    \includegraphics[]{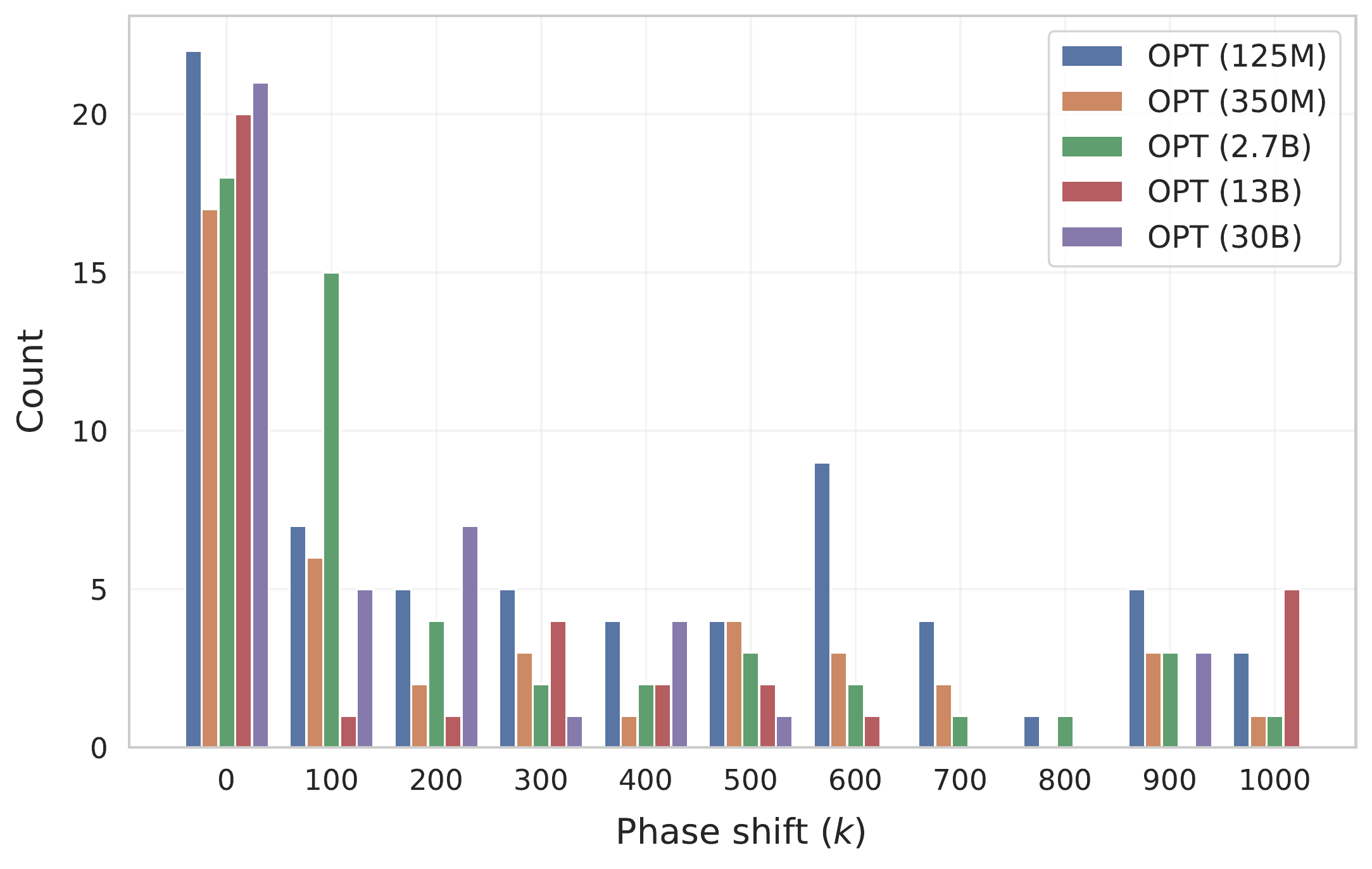}}
    \caption{Distribution of prompts with best accuracy across all six tasks.}
    \label{fig:prompt_best_ps}
\end{figure}

In most tasks, the performance deteriorates when the model process inputs in any other phase shift than zero, especially in zero-shot inference. 
More importantly, the model's performance is not always adversely affected by phase shifts. In fact, \autoref{fig:prompt_best_ps} shows that non-zero starting positions result in the best accuracy for many prompts.
This erratic performance is present in all model sizes, and scaling the number of parameters does not help.
Furthermore, one can see larger models are more affected by shifted starting position, which suggests that absolute positional embedding might need more data or training as the number of parameters increases.

\section{Impact of phase-shifts on fine-tuning}
\label{sec:finetuning}

\begin{figure}
    \centering
    \resizebox{\linewidth}{!}{
    \includegraphics[width=\textwidth]{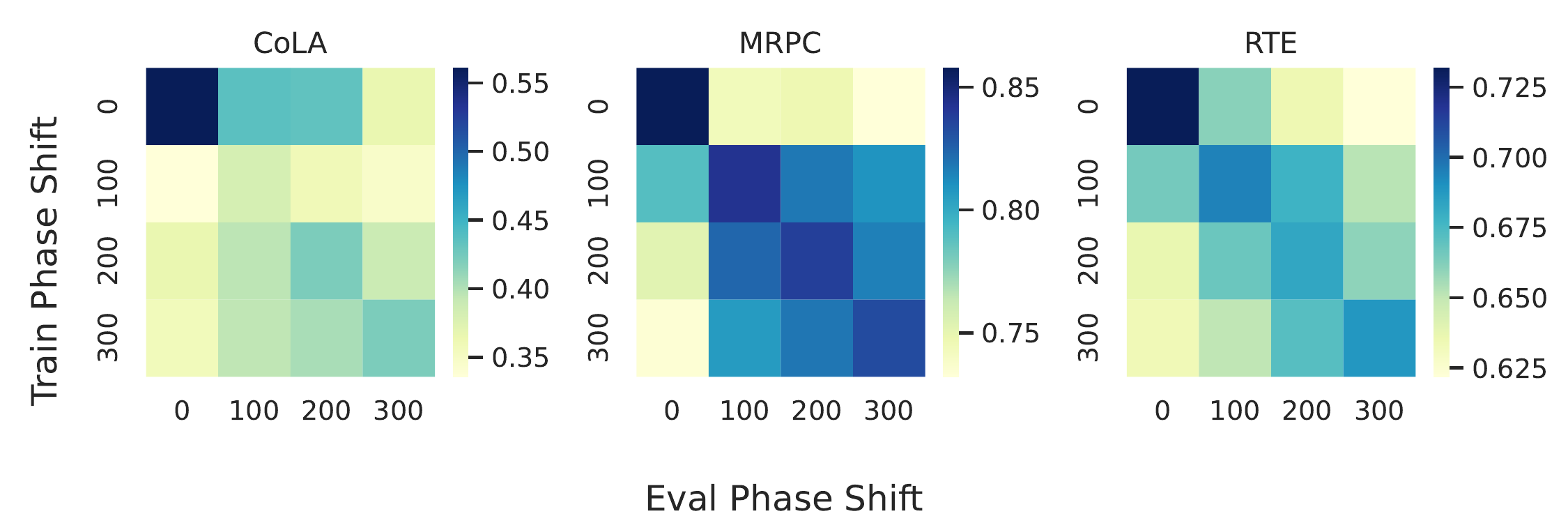}}
    \caption{GLUE task heatmap with varying fine-tuning train and test phase shifts, averaged across all models. Darker colors represent better task performance.}
    \label{fig:glue_phaseshift}
\end{figure}

Finally, we investigate the effect of phase shift in fine-tuning. 
We ask whether the models can generalize to out-of-phase sentences for a given task. 
We train RoBERTa, BART, GPT2 and OPT models on CoLA, RTE and MRPC tasks from the GLUE benchmark \citep{Wang2018:GLUE} and evaluate them on phase-shifts. 
We choose these three relatively small tasks in order to decrease the number of gradient updates to position embeddings during fine-tuning. 
We perform a cross-phase analysis by training and evaluating across different phase shifts ($k={0,100,200,300})$ for all models on the same set of datasets, and show the averaged performance. 
We observe for all models, the task performance drops during out-of-phase evaluation (non-diagonals in \autoref{fig:glue_phaseshift}). 

The drop in performance of evaluating out-of-phase sentences might just be simply attributed to overfitting on position information during fine-tuning. 
However, we observe that for all tasks, training and evaluating on the same phase-shift is worse when $k \ne 0$ (diagonals in \autoref{fig:glue_phaseshift}).
Out-of-phase training appears to be worst for CoLA, which suffers drastically when fine-tuning on different phase shifts.
These results highlight a potential task data bias with respect to different positions. 













\section{Conclusion}

In this work, we investigate the abilities of APEs in encoding the relative positions of the tokens in an input. We observe that TLMs using APEs encode sentences differently based on the starting position of the sentence in the context window. This result has major implications in the way we perceive the sentence processing capabilities of TLMs. 
Specifically, we observe that the representation of the same sentence varies depending on where it is in the context window, such that it impacts zero shot, few shot and full shot task performance of sub-window sentences. 
Future work could leverage the start position in building robust and position-generalizable models. We hope our work can inform the community on the pitfalls of using APEs, and inspire development and adoption of alternative relative position embedding based approaches.

\section*{Limitations}

Our work primarily focuses on evaluating the relative position encoding of APEs. We do not focus on the relative position embeddings \cite{shaw-etal-2018-self, Raffel2020:T5} (RPE) as our method of phase-shift analysis is not applicable to those classes of models. RPEs employ a window based position information computation on the fly, which does not require it to store embeddings uniquely for each position. Thus, a phase shift in RPE would not change the sentence processing pipeline, as the model recomputes the position information based on the shifted window. Thus, we need different tools to study the relative position encoding of RPE than the one proposed in this paper.

We also acknowledge that our study is primarily focused on English language data from BLiMP and GLUE. It is likely the same results would hold in a multi-lingual model, however, since many languages are less word order inflexible than English, that should be investigated in a follow-up work.

\section*{Ethical Consideration}

Our work aims at understanding the difference in sentence representation by shifting position information. In practice, this could yield un-intended results from a TLM deployed in production. Since we observe a large variation in results, we would advise for caution when deploying TLMs in sensitive real world applications, as the relative positioning of a given sentence might evoke different responses from the model. We hope our work can be useful to motivate the use of better positional encoding schemes in pre-training TLMs in future.

\section*{Acknowledgements}

We would like to thank Kanishka Misra, Shagun Sodhani, Stephen Roller and Kushal Arora for their feedback on the initial versions of this draft. 
We are also grateful for anonymous reviewers' feedback.
Siva Reddy acknowledges the support by the Facebook CIFAR AI Chair program.

\bibliography{arxiv}
\bibliographystyle{acl_natbib}
\newpage
\appendix
\modeldetail{!ht}










\section{Experiment Details}
\label{app:exp_details}
\subsection{Models}

We used 11 publicly available pretrained language models in this work, ranging across different architecture families: Encoder, Sequence-to-Sequence, and Auto regressive models. All of them use absolute positional embeddings (APE) that is learned during pretraining. In \autoref{sec:prompting}, we follow the standard practice for in-context learning evaluation \citep{Brown2020:GPT3,Black2022:GPTNeoX, lm-eval-harness} and use autoregressive models. In our initial experiments, we found GPT2 to have a similar behaviour to OPT models, and since the OPT models are available in a wider range of sizes, we primarily focus on them for these experiments.
In fine-tuning (\autoref{sec:finetuning}) and acceptability (\autoref{sec:acceptability}) experiments, we assess all model families. However, because of the computational costs associated with these experiments, we opt for model variants with \textless~1B parameters. The details of all models can be found in \autoref{tab:model_detail}. We use HuggingFace \citep{huggingface} model hub to load, fine-tune train, and run infererence for all models.

\subsection{Datasets}
\label{app:dataset}
We use BLiMP \citep{warstadt-etal-2020-BLiMP-benchmark} for the grammatical acceptability experiments in \autoref{sec:acceptability} as it is typically employed in a inference-only setting and does not require additional training. For \autoref{sec:finetuning}, we take three tasks from the standard language understanding benchmark GLUE \citep{Wang2018:GLUE} which is often used for finetuning language models: MRPC, RTE, and COLA. In addition to these three tasks, we use four other datasets, COPA, PIQA, WinoGrande, and ARC, on which the OPT family have previously demonstrated good performance \cite{Zhang2022:OPT}. \autoref{tab:dataset_detail} shows the statistics of all datasets, and the following provides a brief description of them:

\begin{itemize}
    \item \textbf{BLiMP} \citep{warstadt-etal-2020-BLiMP-benchmark}
    is a challenge set designed to measures the model's ability to distinguish between acceptable and unacceptable English sentences. This benchmark consists of synthetic examples created based on expert-crafted grammars, where each instance comes with two versions: one acceptable and one unacceptable. 
    \item \textbf{COPA} \citep{Gordon2012:COPA} is an open-domain commonsense causal reasoning task, where the model is given a premise and must correctly identify its cause or effect. COPA consists of short hand-crafted sentences and is provided as a multi-choice task.
    \item  \textbf{PIQA} \citep{Bisk2020:PIQA} is a physical commonsense benchmark dataset, challenging language models' idea of the physical world. Given a physical goal, a model must choose the most plausible solution between two choices. This benchmark is used in the multi-choice format. 
    \item \textbf{WinoGrande} \citep{Sakaguchi2020:WINOGRANDE} is a commonsense reasoning benchmark based on the Winograd Schema Challenge (WSC) \citep{Levesque2011:WSC} with increased hardness and scale. The dataset is provided as a pronoun resolution problem, where the model must recover an ambiguous pronoun in a given context.
    
    \item \textbf{ARC} \citep{Clark2018:ARC}  is collected from grade-school-level science questions commonly asked in exams. This question-answering dataset is provided in a multi-choice QA format suitable for evaluating pretrained language models. We use the "\texttt{easy}" subset of this benchmark.
    
    \item \textbf{MRPC} \citep{Dolan2005:MRPC} is a paraphrase identification dataset collected from online news websites and has become a standard benchmark in the NLP community. We follow the previous works and treat the data as a text classification task.
    \item \textbf{RTE} \citep{Dagan2005:RTE} is one of original subtasks in the GLUE benchmark and comprises textual entailment challenges. We follow the standard format and use Natural Language Inference (NLI) protocol for this dataset.
    
    \item \textbf{CoLA} \citep{warstadt2018:CoLA} is a linguistic acceptability dataset, where each example is an English sentence annotated with a binary label showing whether it is a grammatical sentence. This is a text classification dataset and we follow the standard protocol and report Matthews correlation coefficient \citep{Matthews1975:mcc}.

\end{itemize}

\datasetdetail{!h}

\subsection{Grammatical acceptability}
We use all 67 subsets (a total of 67K data instances) of BLiMP \cite{warstadt-etal-2020-BLiMP-benchmark}. A model achieves a score of 1 if it successfully assigns a lower perplexity to the grammatical version of each example. We report the average score across the entire dataset for starting positions that are shifted in the intervals of 10.
The inputs are fed to the models in the format explained in \autoref{app:trigger_tokens}.
Recall that perplexities are ill-defined in case of Masked Language Models. Thus, we follow the formulation of \citet{salazar2019masked} to compute a pseudo-perplexity for RoBERTa and BART.
We adopt the Minicons \cite{misra2022minicons} library to compute the perplexities, which provides a unified interface for models hosted in HuggingFace \citep{huggingface}.

\fthyperparam{!ht}

\subsection{Prompting}
\label{app:prompting}
\begin{table*}
\centering
\footnotesize
\resizebox{\textwidth}{!}{%
\begin{tabular}{
    l@{\hskip 0.3in}
    l@{\hskip 0.3in} 
    l
}
\toprule 
Dataset &  \multicolumn{2}{c}{Template}    \\
\midrule

COPA & 
Prompt &
\texttt{<premise>} because/therefore \texttt{<possible-continuation>}
\\[1mm] 
& Example &\emph{The water in the teapot started to boil} therefore \emph{the teapot whistled.}
\\
\midrule
PIQA & 
Prompt &
 \makecell[l]{Question: \texttt{<question>}$\backslash$ n \\   Answer: \texttt{<possible-answer>}}
\\[3mm] 
& Example & \makecell[l]{Question: \emph{How can I quickly clean my blender without washing?}$\backslash$ n\\
Answer: \emph{Put some ice, water, and a half cup of baking soda in the blender and puree for 3 min.}}\\

\midrule
WinoGrande & 
Prompt &
 \makecell[l]{\texttt{<context>} because \texttt{<replaced-pronoun>} \texttt{<continuation>}} 
\\[1mm] 
& Example & \makecell[l]{\emph{Angela was better suited to conduct the science experiment than Katrina} because \emph{Katrina was less disciplined.}}\\

\midrule
ARC & 
Prompt &
 \makecell[l]{Question: \texttt{<question>}$\backslash$ n \\   Answer: \texttt{<possible-answer>}}
\\[3mm] 
& Example & \makecell[l]{Question: \emph{Amanda is learning about different adaptations of animals. Which is an example of a behavioral adaptation?}$\backslash$ n\\
Answer: \emph{migration of songbirds}}\\

\midrule
MRPC & 
Prompt &
 \makecell[l]{Sentence 1: \texttt{<sentence1>}$\backslash$ n \\Sentence 2: \texttt{<sentence2>}$\backslash$ n \\ Question: Do both sentences mean the same thing?$\backslash$ n \\Answer: \texttt{<label>}}
\\[6.5mm]  
& Example & \makecell[l]{Sentence 1: \emph{Inamed shares closed down nearly 12 percent on Nasdaq, where it was one of the top percentage losers.}$\backslash$ n \\Sentence 2: \emph{Inamed shares dropped as much as about 16 percent on Nasdaq, where it was one of the top percentage losers.}$\backslash$ n \\ Question: Do both sentences mean the same thing?$\backslash$ n \\Answer: \emph{yes}}
\\

\midrule
RTE & 
Prompt &
 \makecell[l]{\texttt{<premise>}$\backslash$ n \\Question:  \texttt{<sentence2>}. True or False?$\backslash$ n \\Answer: \texttt{<label>}}
\\[5mm]
& Example & 
\makecell[l]{\emph{United States astronaut Sunita Williams, currently on board the International Space Station, has today broken the record for\dots}$\backslash$ n \\Question: \emph{Anousheh Ansari paid to go in space}. True or False?$\backslash$ n \\Answer: \emph{False}}
\\

\midrule
CoLA & 
Prompt &
 \makecell[l]{\texttt{<sentence>}$\backslash$ n \\Question: Does this sentence make sense?$\backslash$ n \\Answer: \texttt{<label>}}
\\[5mm]
& Example & 
\makecell[l]{\emph{Brandon read every book that Megan did.}$\backslash$ n \\Question: Does this sentence make sense?$\backslash$ n \\Answer: \emph{yes}}\\

\bottomrule
\end{tabular}
}
\caption{Prompt templates used in EleutherAI Language Model Evaluation Harness library \citep{lm-eval-harness}}
\label{tab:prompt_template}
\end{table*}{}

For evaluating zero-shot inference and in-context learning, we make use of EleutherAI Language Model Evaluation Harness \citep{lm-eval-harness}, an open-source library that is used for evaluating autoregressive pretrained language models \citep{Black2022:GPTNeoX}.
In the zero-shot setting, each example is converted to a prompt using task-specific templates. Then, the prompt is fed to the language model to elicit the answer. Similarly, in the few-shot setup, a prompt is created from the concatenation of few dataset examples base on the same template and are prepended as a context to validation instances.
In our experiments, we use default templates provided by the EleutherAI Language Model Evaluation Harness, which can be found in \autoref{tab:prompt_template}. The task performance is computed over the validation set of due to the lack of public test sets, except for ARC, where we evaluate the models on the test set.
We set the number of few-shots examples to be five and randomly sample them from the training set of each dataset. We report the few-shot results averaged over five random seeds. Note that feeding inputs to the models still follows the same protocol introduced in \autoref{app:trigger_tokens}.

\subsection{Fine-tuning}

We fine-tune all models on CoLA, RTE and MRPC tasks from the GLUE benchmark on different values of phase shift $k$, and evaluate across all possible phase shifts. Since RoBERTa only supports 512 positions, and maximum sentence length in these datasets amount to 128, we train models upto $k=300$. For each fine-tuning experiment, we first run a hyperparameter sweep varying learning rate ($0.0001, 0.0002, 0.0003)$ and training batch size $(16, 32)$ (amounting to 6 runs) with 6\% warmup steps, similar to the setting by \citet{Liu2019:RoBERTa}. 
We also set the weight decay to zero in order to not harm the existing positional encodings which are not used during training.
\autoref{tab:finetune_hyperparam} summarizes all of the parameters.
Finally, we choose the best hyperparams and repeat the experiment over five different seeds ($42$ to $46$), and present an aggregate over the results. \autoref{tab:sweep_results} lists the outcome of hyperparameters tuning.

\begin{figure}
    \centering
    \resizebox{\linewidth}{!}{
    \includegraphics[width=\textwidth]{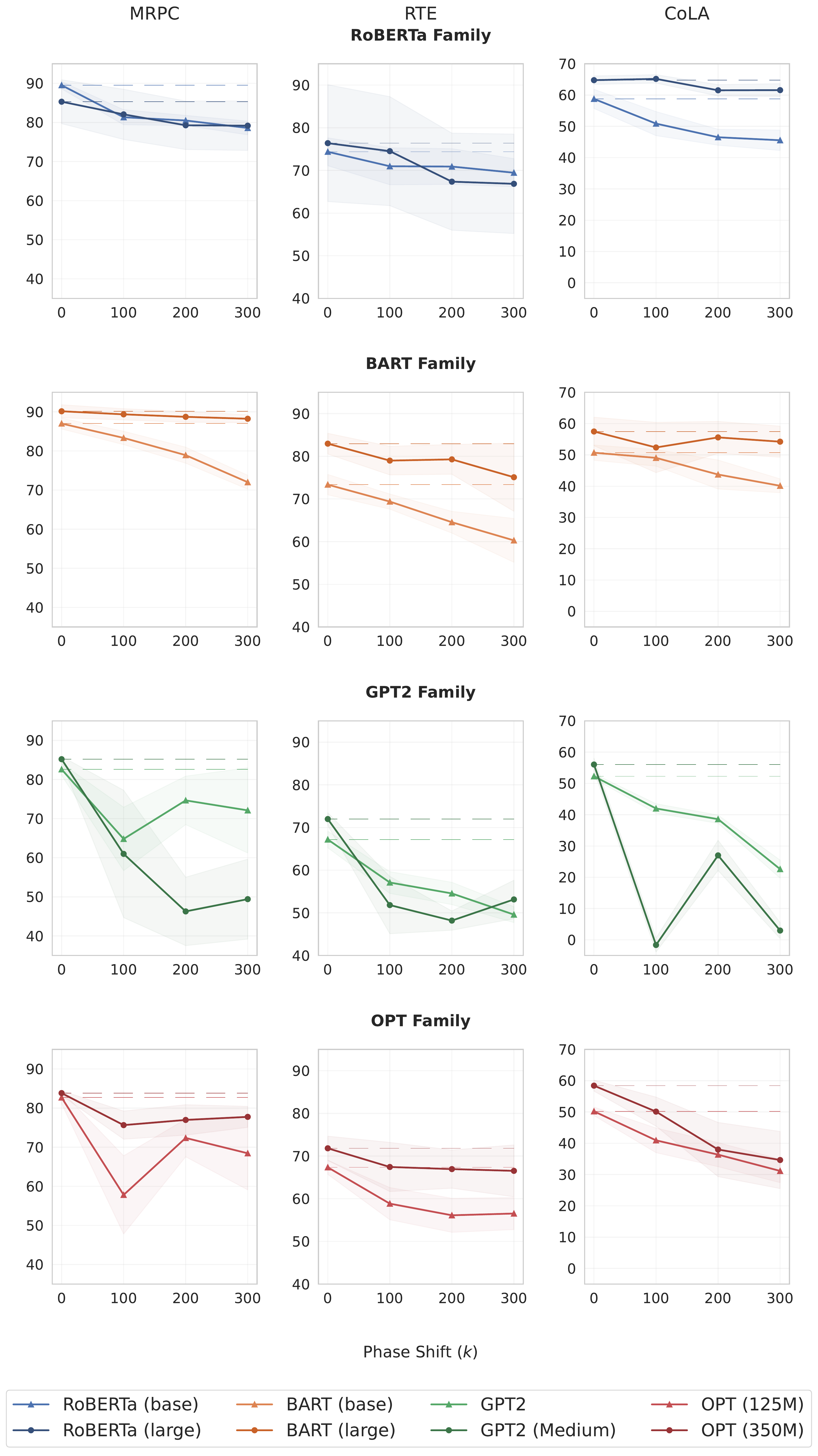}}
    \caption{GLUE downstream task results on CoLA, RTE and MRPC. The dashed lines represent the model performance with no phase shifts. The shaded area show the standard deviation from five random seeds.}
    \label{fig:glue_eval}
\end{figure}

In \autoref{fig:glue_eval}, we further show the difference in fine-tuned models when trained on no phase shift ($k = 0)$ and evaluated on different phase shifts ($k=100,200,300)$. In-line with our experimental results from \autoref{sec:acceptability}, we observe worse generalization results from BART.

\section{Detailed results on phase shifting with prompts}
\label{app:detailed_prompts}
\begin{figure*}
    \centering
    \includegraphics[width=\textwidth]{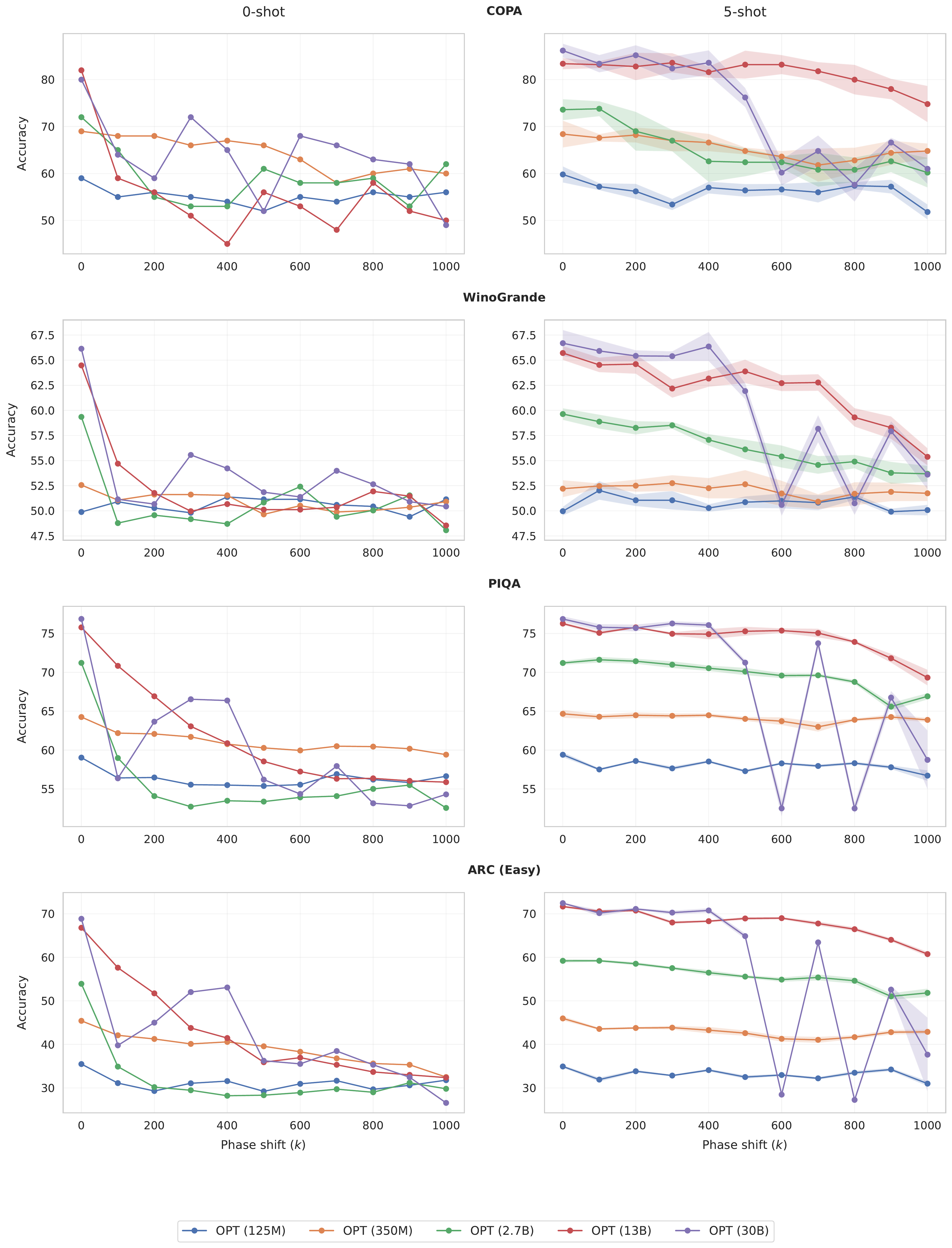}
    \caption{Zero-shot and Few-shot performance of OPT family with various phase shifts for each individual dataset (Part 1)}
    \label{fig:propmt_all_ds_part0}
\end{figure*}

\begin{figure*}
    \centering
    \includegraphics[width=\textwidth]{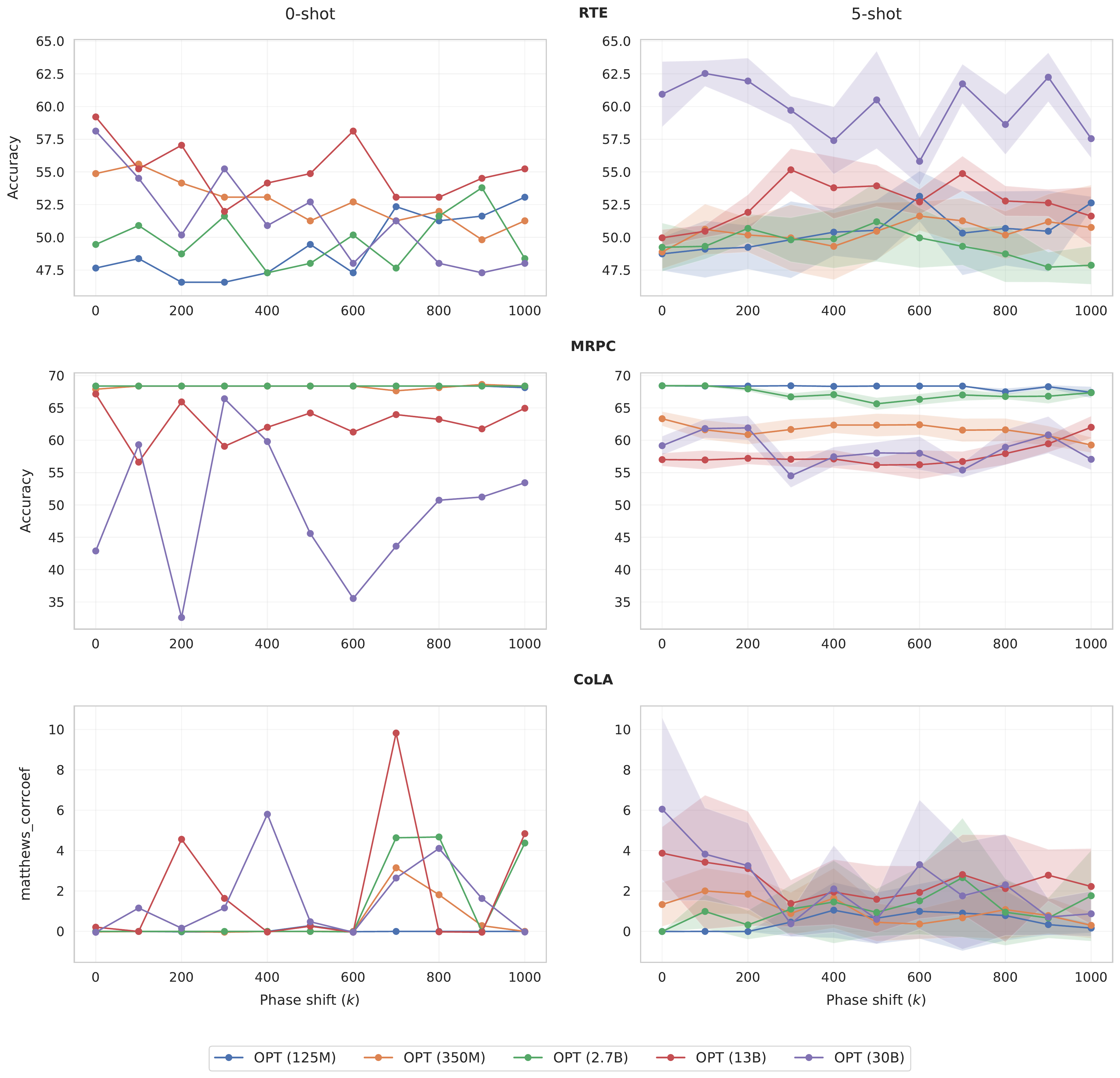}
    \caption{Zero-shot and Few-shot performance of OPT family with various phase shifts for each individual dataset (Part 2)}
    \label{fig:propmt_all_ds_part1}
\end{figure*}

We displayed a holistic view of zero-shot and five-shot experiments in \autoref{fig:prompt_ps_aggr}, covering the accuracies averaged over all six datasets.
In this section, we now report and analyze the result of each dataset individually. \autoref{fig:propmt_all_ds_part0} and \autoref{fig:propmt_all_ds_part1} showcase models' performance in zero-shot and five-shot configurations.
The same pattern can be seen across all model sizes in COPA, WinoGrande, PIQA, ARC (Easy), and RTE. Concretely, the zero-shot abilities of the models sharply decrease as we increase the starting position. 
Moreover, five-shot inference, typically referred to as in-context learning, is also subject to decreased performance, ranging from -2\% to -40\%. However, the degradation is not as severe as with zero-shot setting. 
Only MRPC exhibits stable phase shift performance, but even in this case, larger models are still adversely affected. Due to the exceptionally poor performance of OPT family on CoLA, we exclude these results from our analysis (\autoref{fig:propmt_all_ds_part1}). 

The erratic behaviour observed in majority of evaluated datasets makes it evident that models struggle to encode the relative distances of words as their understanding of inputs heavily change with various phase shifts.
It is important to note that our findings demonstrate models' unstable functioning as opposed to solely highlighting their failure. Indeed, \autoref{fig:prompt_best_ps} shows that one can extract better and improved accuracies with non-zero starting positions.
Namely, $\text{OPT}_\text{30B}$ has the best zero-shot performance on phase shift $k=300$ in the case of MRPC; the same pattern can also be observed in RTE five-shot for $\text{OPT}_\text{13B}$ on phase shift $k=300$.
Another noteworthy observation is that the performance drop is often a \emph{non-monotonic} function of phase shifts. i.e., for some prompts, the model might be more accurate for $k=1000$ than for $k=0$.
This observation suggests that some positional biases might be learned during pre-training and are well-captured by APE. So, increasing values of $k$ in some occasions lands the model attentions in a ``sweet spot'' in the processing window, such that the model benefits from some positional biases learned during pre-training.

We observe the presence of erratic behavior across a fairly wide range of model sizes in the OPT family. Additionally, it can be seen that larger models are more prone to fail at encoding relative positions than their smaller counterparts.
One possible explanation for this is that in order for the models to encode relative positional information, they need to view all combinations of words and sentences in every position. This coverage rarely occurs in natural data, resulting in data sparsity issues. Hence, models with a large number of parameters may require more data/training to learn the relative ordering of words.


\section{Variation of best perplexity across phase shifts}
\label{app:perplexity_phase}

\begin{figure}[!ht]
    \centering
    \resizebox{\linewidth}{!}{
    \includegraphics[width=\textwidth]{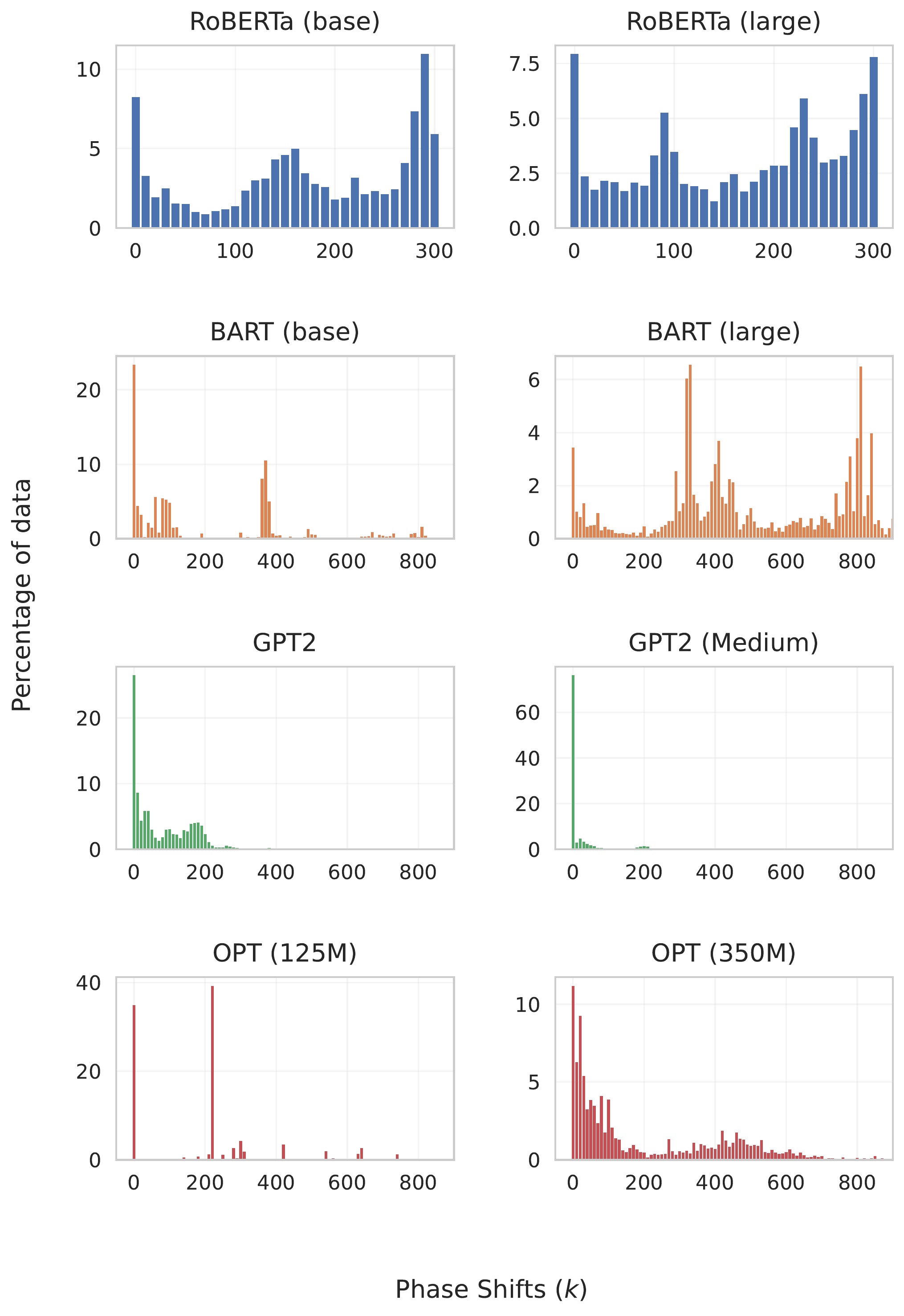}}
    \caption{Distribution of sentences having the lowest perplexities for each phase shift}
    \label{fig:ppl_density}
\end{figure}

In this section, we investigate the perplexity of individual sentences from the BLiMP dataset across each phase shift for each model. We plot the distribution of sentences achieving lowest perplexity in each phase shift for the range of models in \autoref{fig:ppl_density}. We observe several modes of phase shift for RoBERTa and BART models where they have the least perplexity on phase shifts other than the standard (zero position). In the case of GPT2 and OPT, the distribution is more skewed towards zero, indicating they almost always achieve the lowest perplexity in the zero position, i.e. when there is no phase shift.



\section{Code and reproducibility}
For all of the experiments in this work, we used open-source libraries \citep{huggingface,lm-eval-harness,misra2022minicons} and models with publicly available checkpoints. The code to reproduce the results can be accessed from \url{https://github.com/kazemnejad/lm_pos_investigations}.
Furthermore, Listing \autoref{app:code_example} provides a short, easy-to-use code snippet to modify starting position in HuggingFace models. (We will also release a singularity image with all dependencies to facilitate reproducibility.)
We ran our experiments on a mix of NVIDIA A100 40G and NVIDIA RTX8000 48G GPUs.
In particular, almost all experiments required only one of such GPUs. The exception was only in the prompting section, where the $\text{OPT}_\text{30B}$ model required two NVIDIA RTX8000 48G GPUs to fit the model and inputs of batch size 1.

\sweepresults{!ht}
\begin{listing}[!ht]
\begin{minted}[fontsize={\fontsize{7}{10.5}\selectfont},breaklines]{python}
import torch
from transformers import AutoModelForCausalLM, AutoTokenizer

# Download and load the pretrained model
tokenizer = AutoTokenizer.from_pretrained("GPT2-medium")
model = AutoModelForCausalLM.from_pretrained("GPT2-medium")

text = "The capital of France is"
inputs = tokenizer(text, return_tensors="pt")

# Create unshifted position ids from the attention_mask, which is equivalent to torch.arange(inputs["input_ids"].shape[-1])
inputs["position_ids"] = inputs["attention_mask"].cumsum(-1) -1 
print(inputs["position_ids"]) 
# >>> tensor([[0, 1, 2, 3, 4]])

output1 = model(**inputs, return_dict=True)
next_token_id = torch.argmax(output1.logits[0, -1])
print(tokenizer.decode(next_token_id))
# >>>  Paris

# Add special tokens
special_tokens = torch.LongTensor([tokenizer.bos_token_id, tokenizer.eos_token_id])
special_attention_mask = torch.LongTensor([1,1])
inputs['input_ids'] = torch.cat([special_tokens, inputs['input_ids'][0]]).unsqueeze(0)
inputs['attention_mask'] = torch.cat([special_attention_mask, inputs['attention_mask'][0]]).unsqueeze(0)

# Recompute position ids
inputs["position_ids"] = inputs["attention_mask"].cumsum(-1) -1 

# Shift the position ids by 10
inputs["position_ids"] += 9
inputs["position_ids"][0, 0] = 0
print(inputs["position_ids"])
# >>> tensor([[ 0, 10, 11, 12, 13, 14, 15]])

output2 = model(**inputs, return_dict=True)
next_token_id = torch.argmax(output2.logits[0, -1])
print(tokenizer.decode(next_token_id))
# >>>  the

\end{minted}
\caption{Python code example to shift the starting position of a sentence from $k=0$ to $k=10$.}
\label{app:code_example}
\end{listing}

\begin{figure*}
    \centering
    \includegraphics[width=\textwidth]{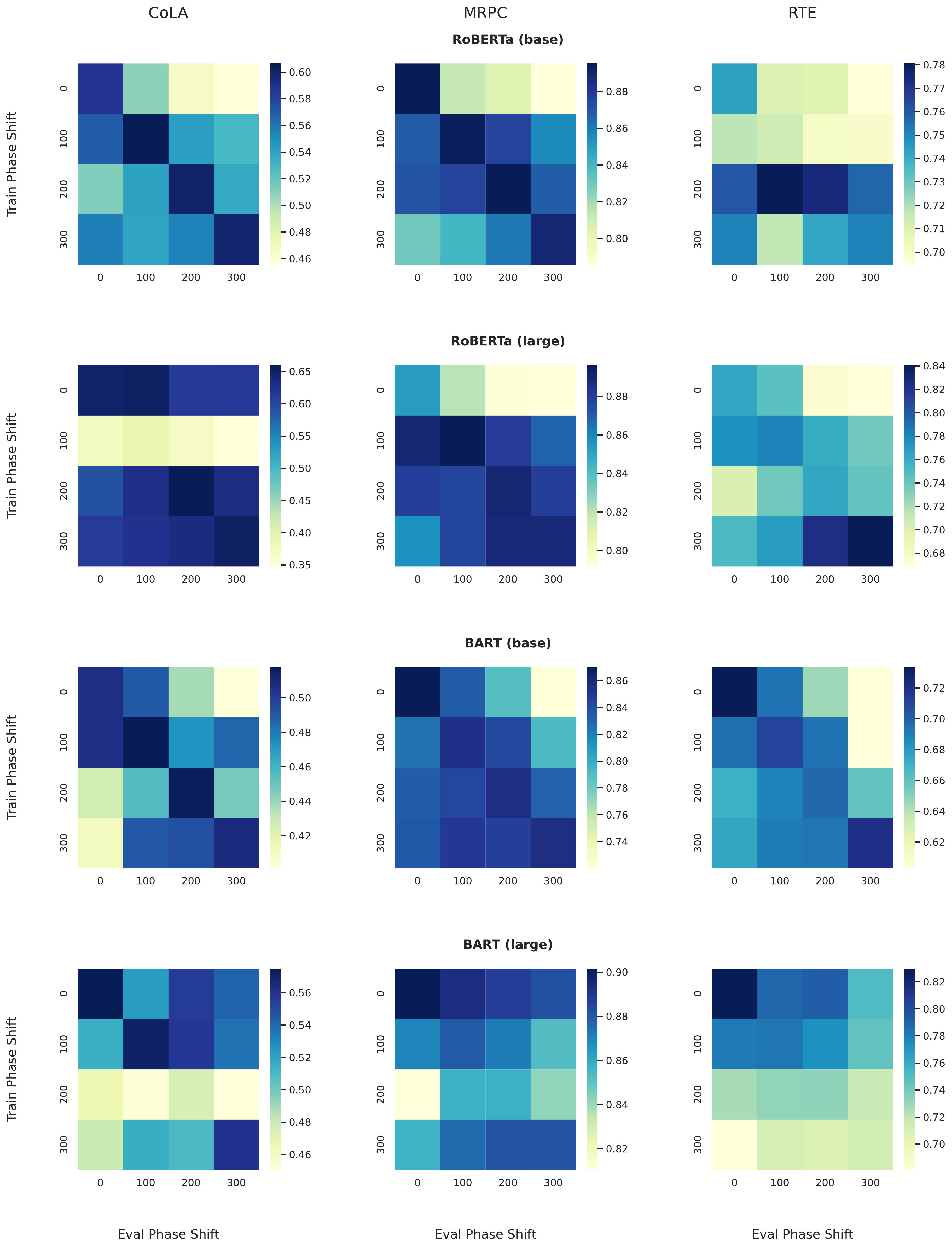}
    \caption{Individual heatmap for each GLUE task and model with varying train (fine-tune) and test phase. (Part 1)}
    \label{fig:heatmap_all_part0}
\end{figure*}

\begin{figure*}
    \centering
    \includegraphics[width=\textwidth]{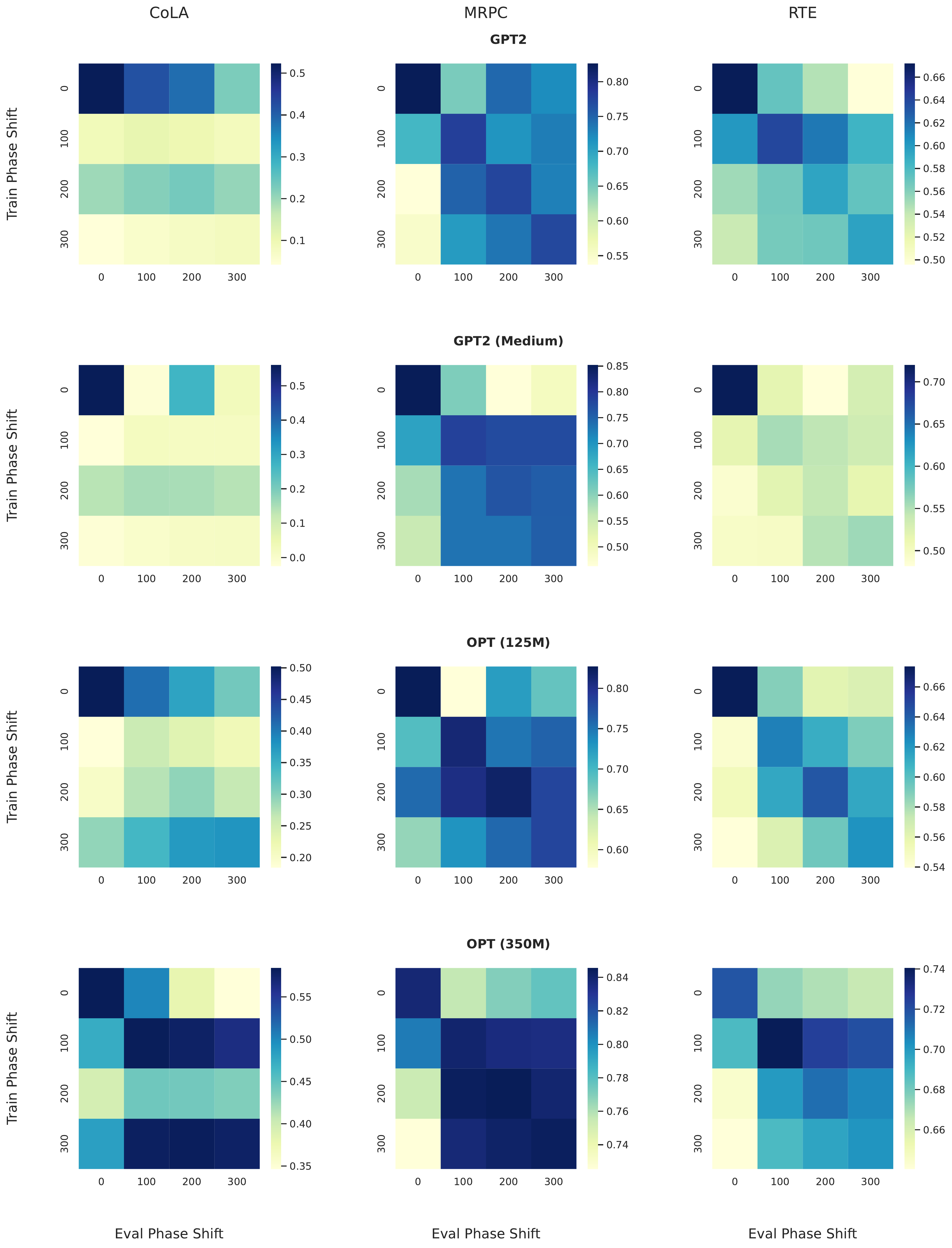}
    \caption{Individual heatmap for each GLUE task and model with varying train (fine-tune) and test phase. (Part 2)}
    \label{fig:heatmap_all_part1}
\end{figure*}

\section{Attention analysis}

We further perform attention analysis on GPT2, RoBERTa and BART to visualize whether the model's attention pattern changes with phase shifts. Following the experimental protocol of \citet{raghu2021vision}, we first collect a summary of attention weights computed with token distances for each token-pair in a sentence. This summary metric is then further normalized for sentence length. The values of this metric show whether the attention is local (low values)---focused on small token distances---or global (high values)---i.e. focused on the whole sentence. 

We compute this attention summary metric on a sample of 5000 sentences drawn from the BLiMP dataset \cite{warstadt-etal-2020-BLiMP-benchmark}. We then plot the summary values per layer and sort according to the values for each attention head, as per \citet{raghu2021vision}. The idea is to discover whether this attention summary metric is drastically different under different phase shift conditions. 

\begin{figure*}
    \centering
    \resizebox{\linewidth}{!}{
    \includegraphics[width=\textwidth]{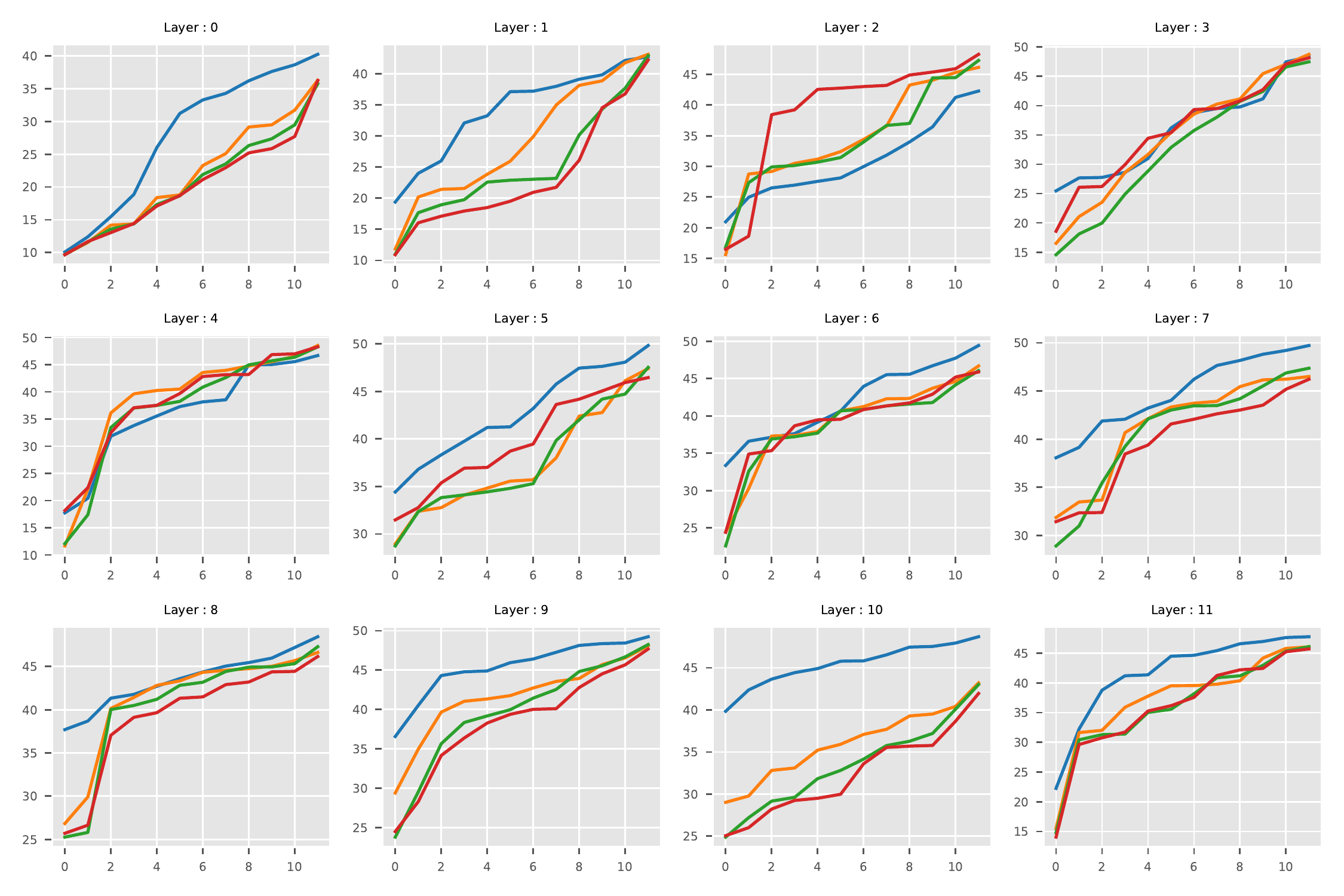}}
    \caption{Attention globality distributions of GPT2 across different heads (sorted according to value) and averaged over all layers and 5000 data points. Blue curve stands for the no phase shift condition, and orange, green and red curves represent $k=100,200$ and $300$ respectively.}
    \label{fig:globality_GPT2}
\end{figure*}

\begin{figure*}
    \centering
    \resizebox{0.8\linewidth}{!}{
    \includegraphics[width=\textwidth]{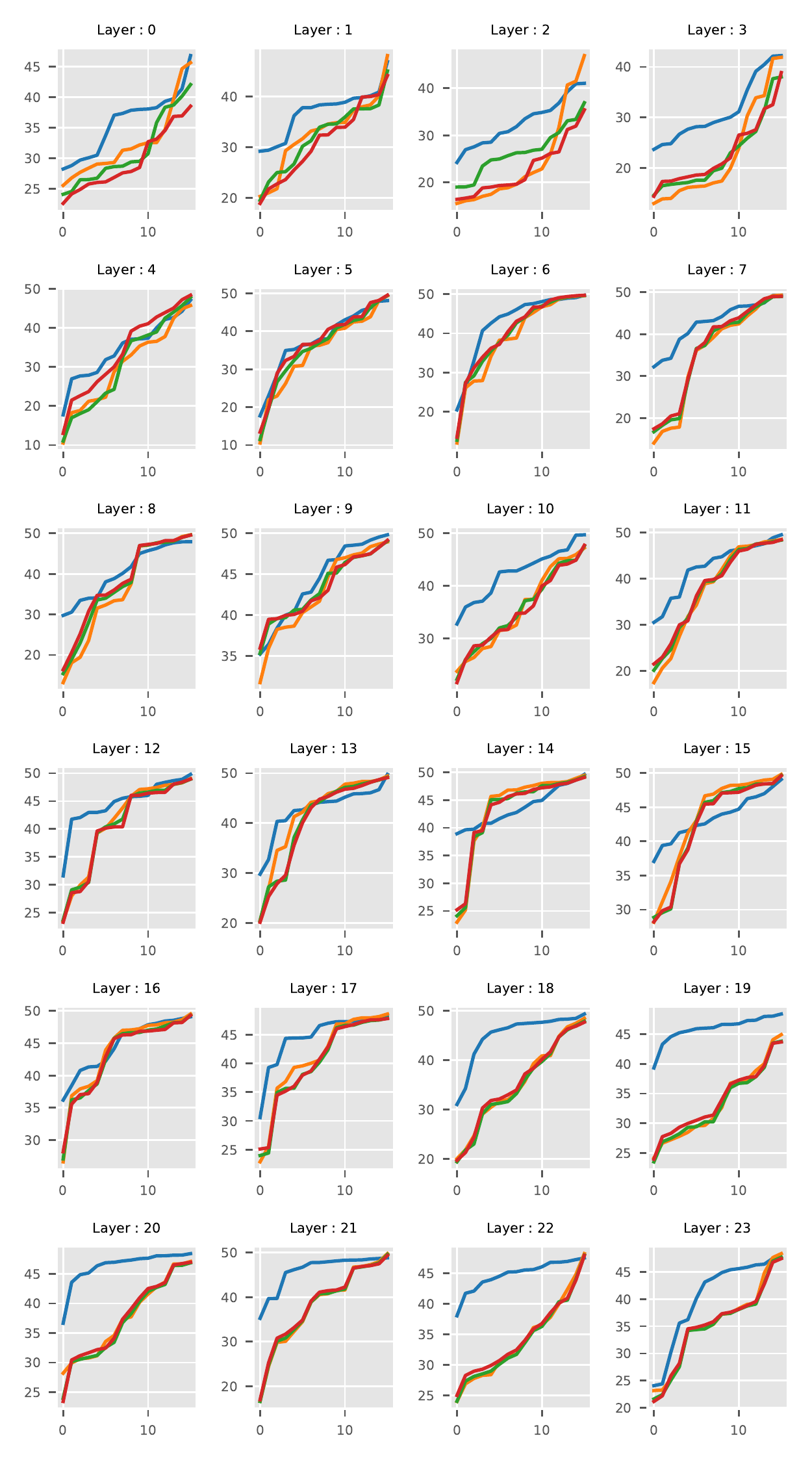}}
    \caption{Attention globality distributions of GPT2-Medium across different heads (sorted according to value) and averaged over all layers and 5000 data points. Blue curve stands for the no phase shift condition, and orange, green and red curves represent $k=100,200$ and $300$ respectively.}
    \label{fig:globality_GPT2-medium}
\end{figure*}

\begin{figure*}
    \centering
    \resizebox{\linewidth}{!}{
    \includegraphics[width=\textwidth]{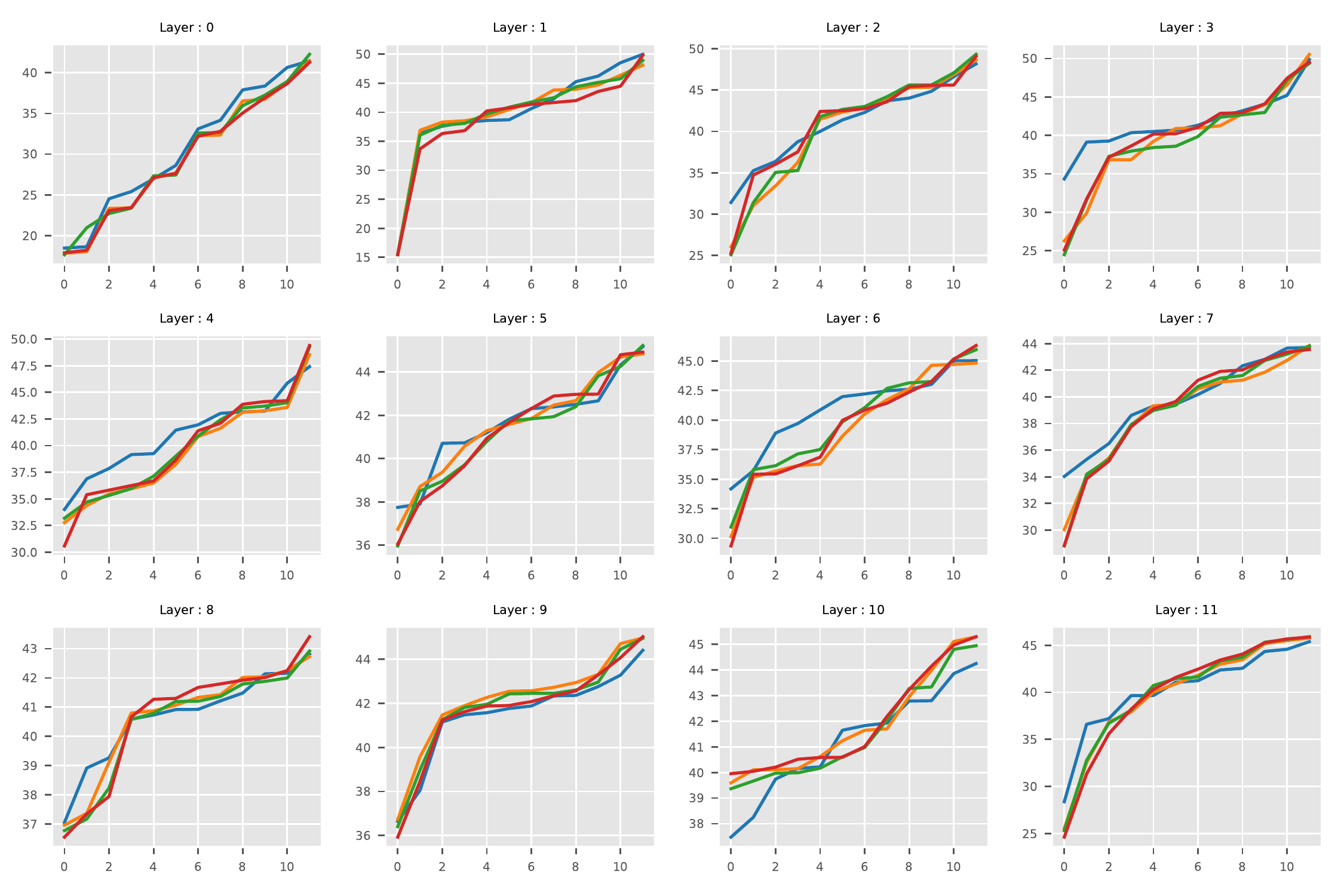}}
    \caption{Attention globality distributions of RoBERTa (base) across different heads (sorted according to value) and averaged over all layers and 5000 data points. Blue curve stands for the no phase shift condition, and orange, green and red curves represent $k=100,200$ and $300$ respectively.}
    \label{fig:globality_roberta-base}
\end{figure*}

\begin{figure*}
    \centering
    \resizebox{0.8\linewidth}{!}{
    \includegraphics[width=\textwidth]{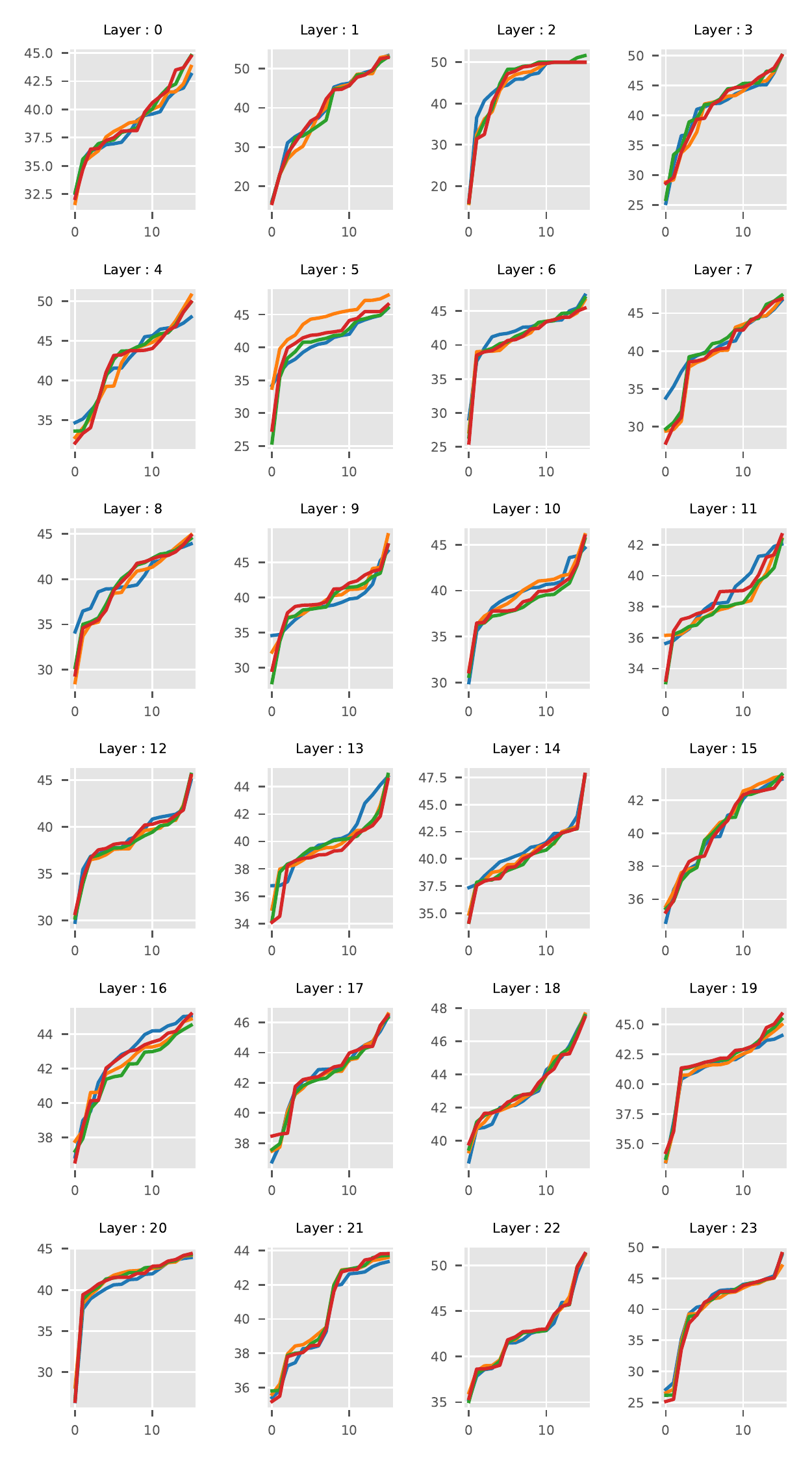}}
    \caption{Attention globality distributions of RoBERTa (large) across different heads (sorted according to value) and averaged over all layers and 5000 data points. Blue curve stands for the no phase shift condition, and orange, green and red curves represent $k=100,200$ and $300$ respectively.}
    \label{fig:globality_roberta-large}
\end{figure*}

\begin{figure*}
    \centering
    \resizebox{\linewidth}{!}{
    \includegraphics[width=\textwidth]{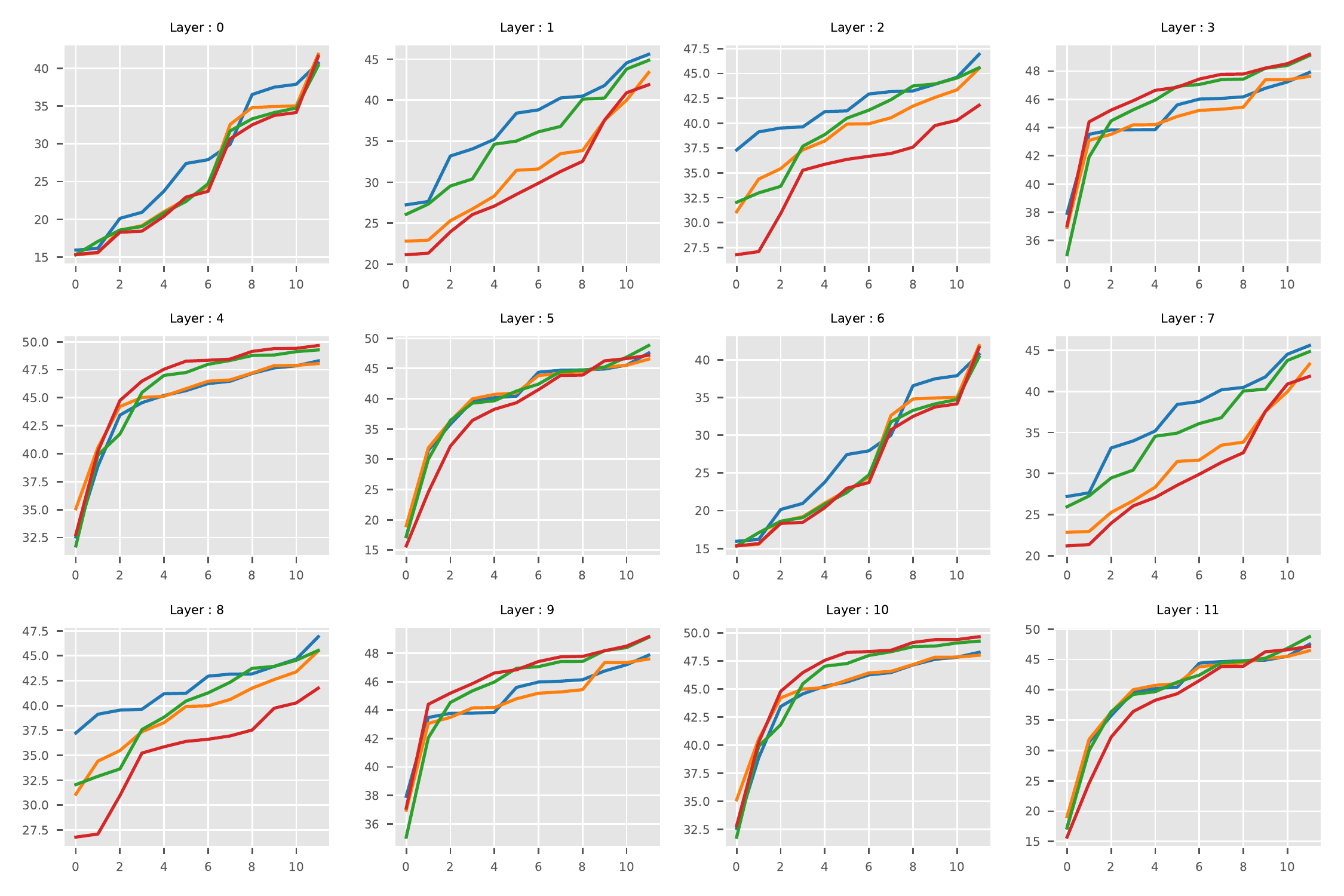}}
    \caption{Attention globality distributions of BART (base) across different heads (sorted according to value) and averaged over all layers and 5000 data points. Blue curve stands for the no phase shift condition, and orange, green and red curves represent $k=100,200$ and $300$ respectively.}
    \label{fig:globality_bart-base}
\end{figure*}

\begin{figure*}
    \centering
    \resizebox{0.8\linewidth}{!}{
    \includegraphics[width=\textwidth]{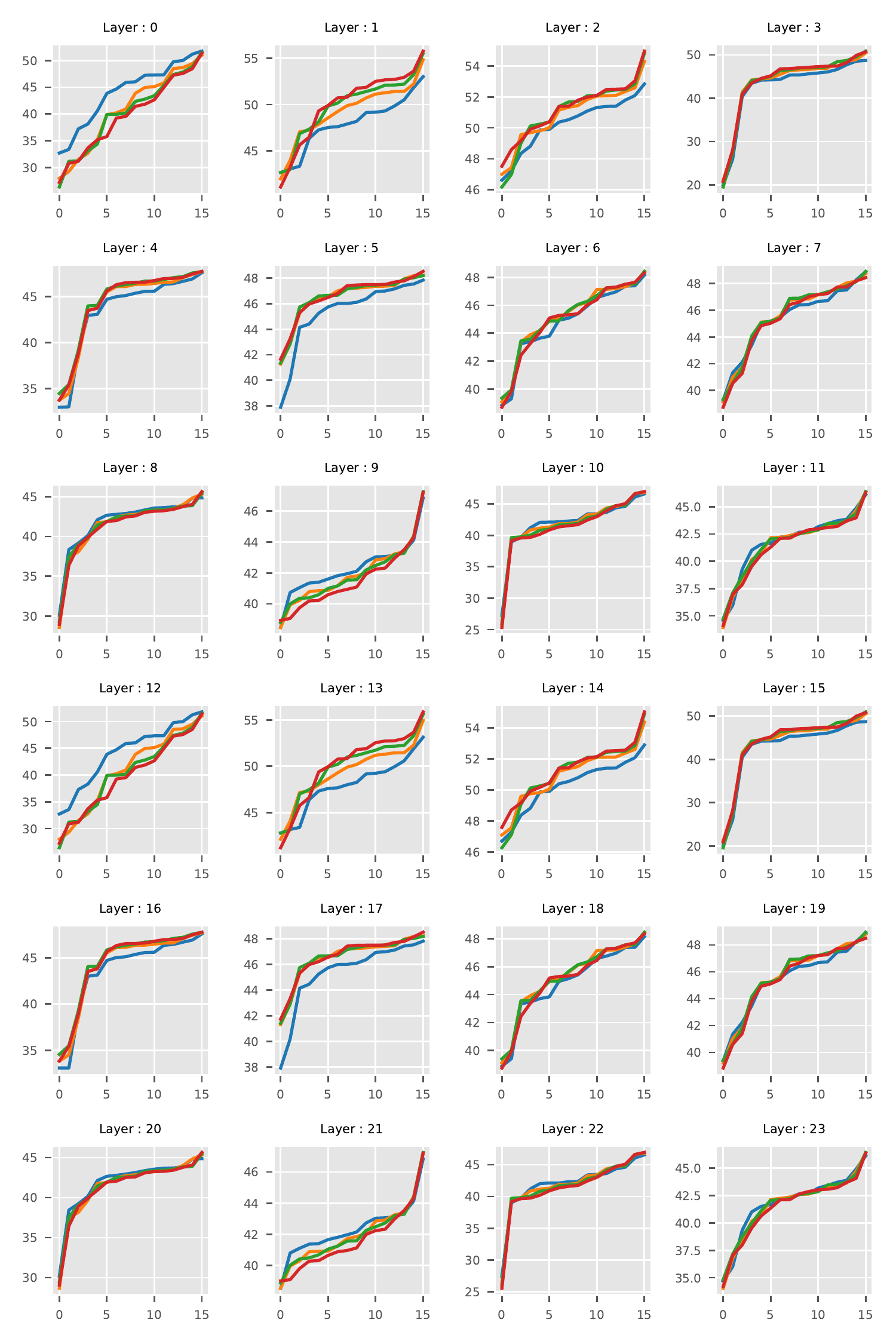}}
    \caption{Attention globality distributions of BART (large) across different heads (sorted according to value) and averaged over all layers and 5000 data points. Blue curve stands for the no phase shift condition, and orange, green and red curves represent $k=100,200$ and $300$ respectively.}
    \label{fig:globality_bart-large}
\end{figure*}

We do observe drastic differences in attention patterns in all layers for GPT2 (\autoref{fig:globality_GPT2}) and GPT2-Medium (\autoref{fig:globality_GPT2-medium}). Comparing this with of RoBERTa (base) (\autoref{fig:globality_roberta-base}) and RoBERTa (large) (\autoref{fig:globality_roberta-large}), we can corroborate our findings from \autoref{sec:acceptability}---RoBERTa is much more robust to phase shifts. Consequently, BART (\autoref{fig:globality_bart-base} and \autoref{fig:globality_bart-large}) also displays differences in attention patterns, but they are not as drastic as GPT2.

\section{Extended Related Work}
\modelposenc{t}
Positional encoding has been always an important part of the Transformer architecture, and since it original introduction different variants of it have been deployed by pretrained models (see \autoref{tab:models_pe} for a summary of positional encoding used by some of popular state-of-the-art models.)

Positional encodings have garnered a niche community over the past several years.
\citet{wang-chen-2020-position} investigate whether position embeddings learn the meaning of positions and how do they affect the learnability for different downstream tasks.
\citet{wang2021on} explore different positional encodings and establish monotonicity, translation and symmetry properties of different methods, including APEs.
They also report that learned APE's demonstrate superior performance for text classification, further adding to the evidence APE's enable exploitation of positional biases.  
\citet{luo-etal-2021-positional} report that masked language model embeddings consists of positional artefacts which bias the model output.
More related to our work, \citet{kiyono2021} train a Transformer model from scratch using shifted positional embeddings for machine translation, and observe improved performance in extrapolation and intrapolation setup.
\citet{haviv2022} reports a surprising finding that autoregressive Transformer models trained without explicit positional information still perform on-par with their counterparts having access to positional information. This result is attributed to the causal attention structure induced by the autoregressive training only, as this effect is not observed with masked language models, as highlighted by both \citet{haviv2022} and \citet{sinha-etal-2021-masked}.
\citet{ke2021} proposes a novel technique to de-correlate the position encodings and token embeddings, and achieve better downstream performance than baselines.
\citet{ravishankar-etal-2021-multilingual} find relative positional encoding does not improve over APE in multi-lingual setting. 

On the other hand, multiple works have shown the advantage of explicit relative positional encoding for length extrapolation.
\citet{csordas2021:devil} show Transformers equipped with variants of relative positional encoding \citep{dai-2019:transformerxl, shaw-etal-2018-self} significantly outperform their absolute counterparts when it comes to length generalization. 
In the same line of work, \citet{ontanon2022:compgen} also find that for numerous synthetic benchmarks, the best extrapolation performance can only be obtained by relative positional encoding.
\citet{press2022train} take the experiments beyond synthetic datasets and show that APE's struggle in generalization to longer sequence of natural language. 
All of these amount to the evidence that points to APE's as one of the potential reasons Transformers are known to fail in length generalization and productivity \citep{pcfg,Lake2018:SCAN}.
Although the benefits of using explicit relative positional bias is mentioned in various works, they typically come at the cost of slowing the training down: \citep{press2022train} report that training T5 (which uses a relative variant of positional encoding) is almost twice as slow as training a model with sinusoidal absolute embedding. Thus, the gained runtime efficiency allows longer training of the APE model, which in turn enables the further extrapolation capabilities. 
These works suggest that we have a lot left to explore about positional encoding and highlight the fact that the consequences of particular choices is still an open field of ongoing research.

\end{document}